\documentclass[opre,nonblindrev]{informs3}

\usepackage{natbib}
\usepackage[ruled]{algorithm}
\usepackage[noend]{algorithmic}
\usepackage{xspace}
\usepackage{latexsym}
\usepackage{amsfonts}
\usepackage{amssymb}
\usepackage{paralist}

 \bibpunct[, ]{(}{)}{,}{a}{}{,}%
 \def\bibfont{\small}%
\allowdisplaybreaks 

\def\eg{\textit{e.g.}\xspace}

\newcommand{\CUT}[1]{{}}

\renewcommand{\paragraph}[1]{\smallskip \noindent {\textsc{#1}}}

\newtheorem{theorem}{Theorem}

\newtheorem{lemma}[theorem]{Lemma}

\MANUSCRIPTNO{N/A} 

\begin{document}


\RUNAUTHOR{Li, Tang, and Zhou}

\RUNTITLE{Towards Distribution-Free Multi-Armed Bandits with Combinatorial Strategies}

\TITLE{Towards Distribution-Free Multi-Armed Bandits with Combinatorial Strategies}

\ARTICLEAUTHORS{%
\AUTHOR{Xiang-Yang~Li}
\AFF{Department of Computer Science, Illinois Institute of Technology}
\AUTHOR{Shaojie Tang}
\AFF{Jindal School of Management, University of Texas at Dallas}
\AUTHOR{Yaqin Zhou}
\AFF{Information Systems and Technology Design Pillar, Singapore University of Technology and Design}
} 

%

\ABSTRACT{
In this paper we study a generalized version of classical multi-armed bandits  (MABs) problem by allowing for arbitrary constraints on constituent bandits at each decision point. The motivation of this study comes from many situations that involve repeatedly making choices subject to arbitrary constraints in an uncertain environment: for instance,  regularly deciding which advertisements to display online in order to gain high click-through-rate without knowing user preferences, or what route to drive  home each day under uncertain weather and traffic conditions. Assume that there are $K$ unknown random variables (RVs), i.e., arms, each evolving as an \emph{i.i.d} stochastic process over time. At each decision epoch, we select a strategy, i.e., a subset of RVs, subject to arbitrary constraints on constituent RVs.
 We then gain a reward that is a  linear combination of observations on selected RVs.
  The performance of prior results for this problem heavily depends on the distribution of strategies generated by corresponding learning policy. For example, if the reward-difference between the best and second best strategy approaches zero, prior result may lead to  arbitrarily large regret.
  Meanwhile, when there are exponential number of possible strategies at each decision point, naive extension of a prior distribution-free policy would cause poor performance in terms of regret, computation and space complexity.
  To this end, we propose an efficient Distribution-Free Learning (DFL) policy that achieves zero regret, regardless of the probability distribution of the resultant strategies.
 Our learning policy has both  $O(K)$ time complexity and $O(K)$ space complexity. In successive generations, we show that even if finding the optimal strategy at each decision point is NP-hard, our policy still allows for  approximated solutions  while retaining near zero-regret.

}

\maketitle

\section{Introduction}
\label{sec:intro}

A multi-armed bandits problem is a basic sequential decision problem defined by a set of strategies against multiple unknown random variables.
 In the simplest form of MAB problems, i.e., single play, a strategy consists of one random variable.
 In the multi-play version, a strategy involves a combination of more than one random variables.
At each time step, a decision maker selects a strategy, and then obtains an observable reward.
The decision maker learns to maximize the total reward obtained in a sequence of decisions through history observation.
MAB problems naturally capture  the fundamental tradeoff between exploration and exploitation in sequential experiments. That is, the decision maker must exploit strategies that did well in the past on one hand, and explore strategies that might have higher gain on the other hand.
 MAB problems now play an important role in online computation under unknown environment, such as pricing and bidding in electronic commerce \citep{babaioff2012dynamic,babaioff2010truthful}, Ad placement on web pages \citep{chervonenkis2013optimization}, source routing in dynamic networks \citep{polychronopoulos1996stochastic}, and opportunistic channel accessing in cognitive radio networks \citep{li2012almost,zhao2008myopic}. Depending on the assumed nature of the reward process,  MAB problems fall into three fundamental categories \citep{bubeck2012regret}: stochastic, adversarial, and Markovian. In this paper, we focus on stochastic bandits.

Despite of many existing results on  multi-play MAB problems against unknown stochastic environment \citep{anantharam1987asymptotically,kalathil2012decentralized,tekin2012online,audibert2009minimax}, their adopted formulations  does not fit those applications that involve numerous or even exponentially large number of candidate strategies at each decision point.
Since the number of possible combinations of selected variables, thus the number of strategies is exponentially large,  if one simply treats each strategy as an arm, the resulting regret bound is exceptional in the number of variables.
In many domains, e.g., networking and communication, where most related combinatorial problems are NP-hard, the aforementioned approaches would be confronted with inefficiency or even failures.
To this end, we aim to explore a more general formulation for \emph{constrained combinatorial bandit problems}.
Given $K$ unknown random variables that are \emph{i.i.d} over time, a
strategy that consists of at most $N$ random variables is selected
under some general constraints at each decision point; all elements of this selected strategy
are revealed after decision, and the corresponding reward  is a linear
combination of these observed values.
The objective is to minimize the upper bound of regret (or maximize the total reward) over time.

For stochastic MAB problems with exponentially large number of combinatorial strategies,
  we explore general approaches to achieve efficient learning in practical.
Herein ``efficient" means low overhead in terms of computation/communication/space complexity caused by the learning process.
When the combinatorial optimization at each decision point is NP-hard, the policy shall be robust enough to admit approximation  algorithms to facilitate the learning process.
Different approximation algorithms, even those with the same approximation ratio, may have different strengths and weaknesses (e.g., in terms of computation/communication/space complexity or implementation manners), and possibly generate varying strategy sets that impact the regret bound. In our design, the resulting upper bound on regret, which is sublinear with time, is only decided by the approximation ratio regardless of which particular algorithm is used.
This nice feature leaves more design space  for wide variety of applications.

Our problem is partially inspired by the problem studied in \citep{gai2012mab}, but we focus on \emph{distribution-free} bound on regret.
The  upper bound on the expected regret of the proposed LLR policy in \citep{gai2012mab}
is \emph{distribution-dependent} \citep{bubeck2012regret},
as it
includes the term of $\frac{1}{\Delta_{\min}}$, the minimum distance between the best static strategy and any other strategy.
In the limit when $\Delta_{\min}$ tends to zero,
its upper bound becomes vacuous.
Moreover, under the premise of unknown environments in MAB problems,
we are usually unaware of the probability distribution of strategies, thus $\Delta_{\min}$.
Therefore, it raises a dilemma when applying distribution-dependent LLR policy to solve combinatorial MAB problems.
To this end, we design a \emph{distribution-free} learning policy that has zero-regret for the linearly combinatorial MAB problem subject to arbitrary constraints. More specifically, the upper bound on regret is a supremum taken over all possible strategies of probability distribution  on $[0,1]$.

In this paper, we present a novel learning
policy, named Distribution-Free Learning (DFL), whose  time and space complexity
are bounded by $O(K)$. By assuming that the combinatorial problem at each decision point can be solved optimally, DFL can achieve distribution-free zero regret for any linear combinatorial MAB problem.
When the underlying combinatorial optimization is NP-hard,
   we propose an efficient learning policy that admits  approximated solution at each decision point  while retaining distribution-free zero regret.
Besides, for completeness of this work, we also derive distribution-dependency regret bounds for
both cases.
Typical applications of the formulation and our
proposed policies are discussed, including auction, shortest path  and
dynamic channel accessing problems. For those applications involving NP-hard
problems, our analysis and results on approximation solutions enable flexible
and efficient implementation  of our proposed policy in practice.
We evaluate our proposed learning policy through extensive simulations. Our simulation results show that our proposed learning policy  outperforms LLR policy in terms of significantly smaller regret.

The remainder of this paper is organized as follows. We first review related works in
Section~\ref{sec:review}.
We give a
formal description of the linearly combinatorial multi-armed bandits
problem in Section~\ref{sec:formulation}. We  present and analyze our
new policy DFL in Section~\ref{sec:policy}. In Section~\ref{sec:beta}, we
give special analysis of our learning policy for the NP-hard combinatorial
optimizations. In Section~\ref{sec:application}, we  present some
applications of our policy. We evaluate our policy with the application of spectrum sensing in Section~\ref{sec:simulation}. We conclude this paper, and discuss limitations as well as future works in Section~\ref{sec:conclusion}.

\section{Related Work}
\label{sec:review}
Depending on the assumed reward process, we have stochastic, adversarial, and Markovian bandits \citep{bubeck2012regret}.
In some literature \citep{liu2013restlessmab}, they roughly divide the works into two categories, non-Bayes bnadits (including stochastic case and adversarial case) and Bayes bandits (Markovian case).
In this paper we adopt the fist classification as \citep{bubeck2012regret}.
In the adversarial bandits, the reward of each arm is nonstochastic, and in the Markovian bandits, each arm is associated with a Markov process of its own state space.

Regrading to each single-play bandit problem  with a specific reward model of the three, we respectively have the following three classical learning policies: the UCB (Upper Confidence bound)-based   algorithm for stochastic case \citep{lai1985ucb}, the Exp3 for adversary case \citep{auer1995gambling}, and the Gittin's indices for the Markovian case \citep{gittins1979bandit}.
A thorough review on stochastic and nonstochastic bandit problems is available in \citep{bubeck2012regret}, while a textbook by Gittin \citep{gittins2011multi} for Markovian bandits.
On the other hand, according to feedback of the observed information on random variables, the bandits problem can fall into categories of full information, semi-bandit, and full bandit.
The decision maker observes value of all random variables in the case of full information,
  and value of these selected random variables in the case of semi-bandit.
While in full bandit, only the instant reward of the selected strategy is fed back.

In this paper, we mainly focus on stochastic bandits where value of selected random variables can be observed.
The simplest form of bandits is single-play bandits where $N=1$ arm is selected among $K$ ones. The analysis of the stochastic bandit is pioneered in the seminal paper of Lai and Robbins \citep{lai1985ucb} where the UCB algorithm is proposed to solve the single play version. Many papers follow  its basic idea to provide improved bounds on regret or simpler upper confidence bound policies for single-play version \citep{agrawal1995sample,auer2002finite}, or extend it to multi-play variants where a fixed number of $N>1$ arms are selected at a time. In \citep{agrawal1995sample}, it proposes a simple sample-mean based method with regret logarithmic uniformly over time, and \citep{auer2002finite} presents variants of Agrawal's work to achieve logarithmic regret in finite time.

All these aforementioned UCB-type policies are distribution-dependent. In
\citep{audibert2009minimax}, Audibert and Bubeck propose a learning
policy called MOSS that has a distribution-free upper bound on regret
with order of $\sqrt{ n\kappa }$. Our work is inspired by MOSS, but considers
a more general   formulation which includes a set of multiple arms
that has to satisfy an arbitrarily given constraint. The MOSS policy is
proposed to solve single-play bandits, and can not be directly used to
solve multi-play version where the exact value of $N$ may be even
unknown. The policy will be highly inefficient if taking each
combination of random variables as an arm, as both the computation and storage costs are exponentially large, e.g., exponential in $N$. Furthermore, for the
NP-hard combinatorial optimization problems, it is too expensive to
find the best strategies by learning all possible strategies. Thus,
our learning policy provides an efficient policy in regret, storage
and computation for such problems.

For the variant with multi-play, Anantharam et al. \citep{anantharam1987asymptotically} firstly consider the problem that exactly $N$ arms are selected simultaneously.
Gai et al. recently extend this version to a more general problem with arbitrary constraints \citep{gai2012mab}.
The model is also relaxed to a linear combination of no more than $N$ arms.
However, the results presented in \citep{gai2012mab} are distribution-dependent, \eg, an arbitrarily small $\Delta_{\min}$ will invalidate the zero-regret result. In this work, we conduct a thorough analysis on both distribution-dependent and -free cases.
In contrast, our learning policy can achieve zero-regret under both distribution-dependent and -free cases.

We note that Chen et al.\citep{chen2013mab} study a similar combinatorial MAB problem  that admit nonlinear reward function under two assumptions. The objective is to minimize a so-called $(\alpha,\beta)$-approximation regret, which is the difference in total expected reward between the $\alpha\beta$ fraction of the expected reward when always playing the optimal fixed arm, and the expected reward of the playing arms output by an assumed oracle that could compute an arm whose expected reward is at least $\alpha$ fraction of the optimum with probability $\beta$. The regret bound achieves distribution free for some reward functions if the two assumptions on the expected reward are satisfied, i.e., monotonicity and bound smoothness.
Our work differs from theirs in several important aspects.
First and for most, our regret analysis covers all forms of linear combinations without any additional assumption. Second, we  analyze the regret bounds for both optimal solution and approximation solution for the NP combinatorial problems.
Third, we discuss various applications to typical network optimization problems.

Some recent works \citep{liu2013restlessmab,tekin2012online} have studied distributed learning among multiple users under the original multi-play model as in \citep{anantharam1987asymptotically}. Though there is no communication overhead,  both of the approaches basically require exponential time in a single learning round. While with communication among multiple users, Kalathil et al. \citep{kalathil2012decentralized} propose an online index-based learning policy that achieves nearly zero-regret.

Recently, the bandits have attracted much attention from researchers in cognitive radio networks.
This line of works starts from single-user play \citep{zhao2008myopic,ahmad2009optimality}, where each channel evolves as independent and identically distributed  Markov processes with good or bad state.
Due to distributed nature of wireless networks as well as limited computation, storage and energy of wireless nodes, efficient distributed implementation  among multiple users then becomes the main focus of policy design \citep{kalathil2012decentralized}, \citep{liu2013restlessmab,anandkumar2010opportunistic,anandkumar2011distributed,gai2011decentralized}. These works basically assume channel quality  evolving with i.i.d stochastic process over time, and a single-hop network setting where conflict happens if any pair of users choose the same channel simultaneously. Under  nonstochastic channel quality, Li et al. \citep{li2012almost} propose an throughput efficient allocation approach with central control. This approach only costs computation and space complexity $O(MN)$ by exploiting dependency among strategies.

\section{Some Motivation Examples}
\label{sec:motivation}
As mentioned in the introduction, our work on this problem
was motivated by many situations that involve repeatedly making choices subject to arbitrary constraints in an uncertain environment.     In this section we introduce several typical applications that may involve exponential number of candidate strategies at each decision point. We will revisit these problems in Section \ref{sec:application}, and  leverage our learning policy introduced in this work to tackle each of them.
\subsection{Online advertisement placement}
    We start with a classical application, Ad placement, from online recommender/advertising systems. Ad placement is the process of deciding which advertisement to display to users based on their individual history.
     Suppose there is a sequence of $N$ advertising spaces, and
     a pool of $K$ available ads (bandit arms), where $K\geq N$. The payoff of an ad is measured by valid click-through-rate, which is decided by user's preference that are unknown in advance.  The payoff of ad $k$ evolves  as an unknown stochastic process $\xi_{k}$ with mean $\mu_{k}$. And the advertiser wants to maximize its social welfare, summed payoff of all advertisements, by displaying personalized  advertisements to users based on their preferences.

\subsection{Stochastic shortest path problem}
Another example is the stochastic shortest path problem.
Consider a network $G=(V,E)$ with a set $V$ of vertices connected by edges of $E$.
A sequence of packets must to be routed from a distinguished vertex, called source, to another distinguished vertex, called destination.
At each time slot a packet is sent along a specific source-destination path by a routing protocol or a decision maker.
Depending on the congestion, each edge in the network may experience dynamic delay which changes over time. The goal is to find a route whose expected delay is minimized among all passible paths.

 \subsection{Dynamic channel accessing in multi-hop cognitive radio networks}
The third application is the dynamic channel accessing in multi-hop cognitive radio networks.
Given a cognitive radio network described by conflict graph $G=(V,E,C)$ with  a set $V=\{v_i|i=1,\dots,N\}$ of $N$ users,
  a set $E$ of edges, and a set $C=\{c_j|j=1,\dots,M\}$ of $M$ channels.
  Conflicts happen if any two adjacent users access the same channel simultaneously.
At each time slot $t$, user $v_i$ has $M$ choices of channels,
  each having data rate drawn from \emph{i.i.d} stochastic process $\xi_{i,j}(t)$ over time with an unknown mean $\mu_{i,j} \in[0,1]$.
Without loss of generality, we assume that the same channel may demonstrate different channel qualities for different users.
For the same channel $c_j$, the random process $\xi_{i,j}(t)$ is independent from $\xi_{i',j}(t)$ if $i \neq i'$.
The objective of the dynamic channel accessing  problem is to find an optimal allocation of channels for users so that the time averaged throughput is maximized. 

\section{Problem Formulation}
\label{sec:formulation}
We consider a time slotted system with $K$ arms/unknown random variables $\xi_k(t)$, $1 \leq k \leq K$, where $t$ is index of time slot.
 We assume that each of the $K$ variables evolves as  an \emph{i.i.d}
 stochastic process $\xi_{k}(t)$ normalized between $[0,1]$ over time
 with mean $\mu_{k}$, which is unknown a priori.
Table~\ref{notations} summarizes the notations used in this paper.

At each time slot $t$, an $N$-dimensional \emph{strategy} vector
$\mathbf{s}_x=\{s_{x,i}|i=1,\dots,N\}$ is selected under some
\emph{policy} from the \emph{feasible strategy set} $F$. By ``feasible"
we mean that each strategy satisfies the underlying constraints
imposed to $F$. For example, in the previous dynamic channel accessing problem, no two adjacent users can access the same channel simultaneously in any feasible solution. Here $s_{x,i}$ is the index of random variables
selected as the $i$th element of strategy $\mathbf{s}_x$.
We use $x=1,\dots,X$ to index strategies of feasible set $F$ in the
decreasing order of average reward \[\lambda_x=\sum_{i=1}^{N}
\mu_{s_{x,i}}\] \eg, $\mathbf{s}_1$ has the largest average reward. Note that a strategy may consist of less than $N$ random
variables, as long as it satisfies the given constraints.
We then set $s_{x,i}=0$ for any empty entry $i$.
Please also note
that the uniformly linear combination of random variables in a strategy
includes the weighted case, as we can easily take the product of each
arm and its weight as a new random variable, and normalize the new
random variable to $[0,1]$.
If the unknown means were known, the static optimal strategy would be
\begin{eqnarray}
   \mathbf{s}_1 = \argmax_{\mathbf{s}_x\in F}\lambda_x =\argmax_{\mathbf{s}_x\in F} \sum_{i=1}^{N} \mu_{s_{x,i}}
\end{eqnarray}
When a strategy $\mathbf{s}_x$ is determined, one observes the value of ${\xi}_{s_{x,i}}(t)$, and then the total reward of strategy $\mathbf{s}_x$ at $t$ is
    \begin{equation}
    R_x(t)= \sum_{s_{x,i} \in \mathbf{s}_x}{\xi}_{s_{x,i}}(t).
 \end{equation}
We evaluate policies using \emph{regret}, which is defined as the
difference between the expected reward  obtained by a fixed optimal strategy $\mathbf{s}_1$, and the expected reward obtained by our policy.
We define $R_x := E[R_x(t)] $.
Let $R_1 =
\lambda_1$ be the expected average reward of the optimal strategy $\mathbf{s}_1$, and $\Delta_x=R_1 - R_x$  be the distance between $\mathbf{s}_1$ and $\mathbf{s}_x$, then the regret of a strategy $\mathbf{s}_x$ over $n$
time slots can be expressed as
\begin{eqnarray}
    \mathfrak{R}(n)= nR_1 - E\biggl[\sum_{t=1}^{n} R_x(t)\biggl]
    = \sum_{x:R_{x} < R_1} \Delta_x E\bigl[T_x(n)\bigl]
\end{eqnarray}
As these random variables  are unknown, and observed after decision,
we have to learn  the reward of each strategy. We denote the estimated value of strategy $\mathbf{s}_x$ at time slot $t$ by \emph{weight}
$W_x(t)=\sum_{s_{x,i} \in \mathbf{s}_x} w_{s_{x,i}}(t)$, where
weight $w_{s_{x,i}}(t)$ is estimated value of random variable
$\xi_{s_{x,i}}(t)$.
\begin{table}[t]\setlength{\tabcolsep}{3pt}
\begin{center}
\caption{Summary of notations}
\label{notations}
\begin{tabular}{c| c}
\hline
Variable & meaning\\
  \hline
    $K$             & number of arms/random variables    \\
    $\xi_{k}$       &random variable (i.e., arm) with index $k$     \\
    $\mu_{k}$       &mean of $\xi_{k}$     \\
    $\tilde{\mu}_{k}$ &observed mean of $\xi_{k}$ up to current time slot     \\
    $m_{k}$         &number of times arm $\xi_{k}$ has been observed so far           \\
    $\mathbf{s}_x$  & the $x^{th}$ strategy in set $F$, $\mathbf{s}_1$ is the optimal strategy.      \\
    $X$  & the maximum index of strategy in set $F$ \\
    $N$             & length of strategy vector  $\mathbf{s}_x $, $N\leq K$    \\
    $\lambda_x$     & mean reward achieved by $\mathbf{s}_x$                \\
    $\Delta_x$      & $=R_1 - R_x=\lambda_1 - \lambda_x$, the distance between $\mathbf{s}_1$ and $\mathbf{s}_x$    \\
    $\Delta_{\min}$ & $\min_{R_x < R_1 }{\Delta_x}$           \\
    $\Delta_{\max}$ &   $\max_{R_x < R_1 }{\Delta_x}$       \\
    $T_x(n)$        & number of times strategy $\mathbf{s}_x$ has been played by time slot $n$ \\
    $W_x(n)$        & weight (estimated reward) of strategy $\mathbf{s}_x$ at time slot $n$                        \\
    $Z_x$           & $=\lambda_1-\Delta_x/2 $                                                 \\
    $W_1$             & $\min_{1\leq t \leq n} W_1(t)$                                    \\
    $\Delta_{\beta, x}$     & $=\lambda_1/\beta - \lambda_{x}$       \\
    $\Delta_{\beta,\min}$     & $\min_{R_{\beta,x} < R_1/{\beta} }{\Delta_{\beta,x}}$           \\
    $\Delta_{\beta,\max}$     & $\max_{R_{\beta,x} < R_1/{\beta} }{\Delta_{\beta,x}}$           \\
    $T_{\beta,x}(n)$    & number of times that strategy $\mathbf{s}_{\beta,x}$ has been played by time slot $n$         \\
    $x_{\beta}$          & index of the worst strategy $\mathbf{s}_y$ with $\lambda_y \ge \lambda_1 / \beta$ \\
    $Z_{\beta,x}$           & $=\lambda_1/\beta-\Delta_{\beta,x}/2 $ \\
  \hline
  \end{tabular}
\end{center}\vspace{-10pt}
\end{table}

When finding the best strategy at each decision point is NP-hard, we introduce
a weaker version of regret, called \emph{$\beta$-regret}, which is defined as the
difference between $nR_1/ {\beta}$ and the reward that obtained by our policy. We say a policy is $\beta$-approximation policy
if and only if it yields zero time averaged $\beta$-regret.
Let $R_{\beta,x}(t)$ be the reward of strategy $\mathbf{s}_{\beta,x}$  generated by  the $\beta$-approximation policy, the $\beta$-regret can be expressed as
\begin{eqnarray}
    \mathfrak {R}_{\beta}(n)&=&nR_1/ {\beta} -E\biggl[\sum_{t=1}^{n}R_{\beta,x}(t)\biggl] \\
    &=& \sum_{\mathbf{s}_{\beta,x}:R_{\beta,x} < R_1/{\beta}} \Delta_{\beta,x} E\bigl[T_{\beta,x}(n)\bigl] \\
    & & + \sum_{\mathbf{s}_{\beta,x}:R_{\beta,x} \ge R_1/{\beta}} \Delta_{\beta,x} E\bigl[T_{\beta,x}(n)\bigl] \\
    &\leq&  \sum_{\mathbf{s}_{\beta,x}:R_{\beta,x} < R_1/{\beta}} \Delta_{\beta,x} E\bigl[T_{\beta,x}(n)\bigl]
\end{eqnarray}

where $T_{\beta,x}(n)$ is the number of times that strategy $\mathbf{s}_{\beta,x}$ has been played by time slot $n$, and $\Delta_{\beta, x} = R_1/\beta - R_{x}$ is the  distance between $R_1/{\beta}$ and mean reward of strategy $\mathbf{s}_x$. Here 
all strategies can be divided into two sets, i.e., a set of $\beta$-approximation strategies and a set of non-$\beta$-approximation strategies.
A $\beta$-approximation strategy is a strategy with mean reward of at
least $R_1/{\beta}$,
and a non-$\beta$-approximation strategy is one with mean reward less than $R_1/{\beta}$.
Thus we have negative $\Delta_{\beta,x}$ for $\beta$-approximation strategies and positive $\Delta_{\beta,x}$ for non-$\beta$-approximation strategies. Hereby let  $\Delta_{\beta,\min}=\min_{R_{\beta,x} < R_1/{\beta} }{\Delta_{\beta,x}}$.



In both  cases, we expect regret $\mathfrak{R}(n)$ (or
$\mathfrak{R}_{\beta}(n)$)  to be as small as possible. Intuitively,
if the regret is $o(n)$, sublinear with time $n$, then the time
averaged regret will  approach $0$, indicating time averaged reward to
be maximum.
Though some existing learning policies can achieve zero-regret, their regret bound heavily depends on the distribution of strategies in
feasible set. That is, the upper bound of regret
$\mathfrak{R}$ (resp. $\beta$-regret $\mathfrak {R_{\beta}}$) including a
factor of $\frac{1}{\Delta_{\min}}$ (resp. $\frac{1}{\Delta_{\beta,\min}}$)
that becomes vacuous if $\Delta_{\min}$ (resp. $\Delta_{\beta,\min}$)
$\rightarrow 0$.
To this end, we aim to design a zero regret (resp. $\beta$-regret) policy without dependency on $\Delta_{\min}$ (resp. $\Delta_{\beta,\min}$).

\section{Distribution-free Learning Policy}
\label{sec:policy}

\subsection{Naive method}

A naive method for a distribution-free policy of our combinatorial
NP-hard MAB problem is to treat each strategy  $\mathbf{s}_x \in F$ as an
arm, by which we can directly use the MOSS policy to achieve the following regret
without $\Delta_{\min}$.

\begin{theorem}
    \label{theorem-moss}
\cite{audibert2009minimax} MOSS satisfies
$\sup \mathfrak{R}(n) \leq 49 \sqrt{n \kappa},$
where the supremum is taken over all $\kappa$-tuple of probability
distributions on $[0, 1]$.
\end{theorem}

Here $\kappa$ is actually the number of strategies available (i.e.,
$\kappa\simeq \Theta(K^N)$ for combinatorial strategies) as MOSS
is proposed for single-play bandit.  MOSS yields regret growing
linearly with the square root of the number of strategies, which is
inefficient when the feasible strategy set $F$ has
exponentially large number of unknown strategies. Meanwhile, it leads to extremely high computation and storage costs, which is exponential in $N$, for updating and storing observed information of all strategies. Thus the naive approach has poor performance in terms of regret, computation and space complexity.
When the combinatorial problem is NP-hard, it does not admit efficient
approximation algorithms on strategy decision as well. To resolve the above issues, we introduce a novel learning policy in the rest of this paper.

\subsection{Distribution-free Learning Policy (DFL) policy}
In this section, we  present a novel  policy, called DFL, that is  a distribution-free
zero-regret learning policy for combinatorial strategies (described in
Algorithm~\ref{alg-dfl})
with low cost to store and update observed information by exploiting
dependencies among correlated strategies.

\begin{algorithm}[ht]
\caption{Learning policy DFL}
\label{alg-dfl}
  \begin{algorithmic}[1]
   \STATE For each round $t=0,1,\dots,n$ \\*
    Select a strategy $\mathbf{s}_x$ by maximizing
        \begin{equation}
        \label{e-max}
          \max_{\mathbf{s}_x\in F} \sum_{s_{x,i} \in \mathbf{s}_x}
             \biggl( \tilde{\mu}_{s_{x,i}}(t)
                    +\sqrt{
                         \frac{ \max{(\ln{\frac{t^{2/3}}{K m_{s_{x,i}}} }},0)} {m_{s_{x,i}}}
                           }
             \biggl)
        \end{equation}
  \end{algorithmic}
\end{algorithm}


For brevity, let weight
\begin{equation}
 w_{s_{x,i}}(t+1)= \tilde{\mu}_{s_{x,i}}(t)  +\sqrt{\frac{\max{(\ln{\frac{t^{2/3}}{K m_{s_{x,i}}} }},0)} {m_{s_{x,i}}}}
\end{equation}

    be estimated reward of $\xi_{s_{x,i}}(t+1)$ and weight
\begin{equation}
    W_{x}(t+1)=\sum_{s_{x,i}\in \mathbf{s}_x} w_{s_{x,i}}(t+1)
\end{equation}

denote estimated reward of strategy $\mathbf{s}_x$.
As shown in Algorithm~\ref{alg-dfl}, our proposed learning policy requires storage linear with $K$ to update observed reward.
\begin{theorem}
    \label{theorem-complexity}
    Algorithm 1 has time and space complexity of $O(K)$, even though the number of strategies may grow exponentially to $\Theta(K^N)$.
\end{theorem}

Here we have assumed that we can instantly find a strategy with maximum reward in (\ref{e-max}). In Section \ref{sec:beta}, we further show that even if finding such a strategy is NP-hard, our policy still allows for  approximated solutions  while retaining zero-$\beta$-regret.
Below we give the main results on the regret bound of Algorithm \ref{alg-dfl}.

\begin{lemma}
    \label{lemma-opt-free}
    The regret of  policy DFL satisfies
    \setlength\arraycolsep{2pt}
      \begin{eqnarray}
            \sup \mathfrak{R}(n) &\leq &
  \label{suprn}                 
            NK +
               \sqrt{ K e} n^{\frac{2}{3}} +
               16N^3 n^{3/4} +
              \biggl[\frac{K}{e^2}+
              (1+  4 \sqrt{K}N^2 )N
              \biggl]NK n^{\frac{5}{6}}
      \end{eqnarray}
      without dependency on $\Delta_{\min}$. The supremum is taken over all $X$-tuple of probability distributions on $[0,1]$.
\end{lemma}
\emph{Proof:} See Appendix.

For completeness of the paper, we also derive the following regret bound with dependency on $\Delta_{\min}$. When $\Delta_{\min}$ is far beyond zero, this may provide a tighter regret bound.

\begin{lemma}
    \label{lemma-opt-depend}
     DFL has distribution-dependent regret
    \setlength\arraycolsep{2pt}
      \begin{eqnarray}
        \mathfrak{R}(n)  \leq
        \frac{e^3 K^3}{\Delta_{\min}^5}
                            +
                            NK \biggl(
                           1 + \frac{16N^2  \ln{(\frac{n^{2/3}}{K} N^2)}}{\Delta_{\min}^2}
                            + \frac{K n^{1/3}}{e^2}
                            +
                           \frac{8 N^3 K \ln{(\frac{n}{K} N^2)}}{\Delta_{\min}^2} {n^{\frac{1}{3}}}
                                        + \frac{KN}{(1-1/e)\Delta_{\min}^2}
                           \biggl)
      \end{eqnarray}
\end{lemma}
\emph{Proof:}
 See Appendix.

Lemma \ref{lemma-opt-free} and \ref{lemma-opt-depend} together imply the following main theorem.
\begin{theorem}
The regret of  DFL is bounded by
    \begin{eqnarray}
       \mathfrak{R}(n) \leq \min \biggl\{
          NK +
               \sqrt{ K e} n^{\frac{2}{3}} +
               16N^3 n^{3/4} +
              \biggl[\frac{K}{e^2}+  
              (1+  4 \sqrt{K}N^2 )N
              \biggl]NK n^{\frac{5}{6}}
        ,
        \nonumber\\
        \frac{e^3 K^3}{\Delta_{\min}^5}
                            +
                            NK \biggl(
                           1 + \frac{16N^2  \ln{(\frac{n^{2/3}}{K} N^2)}}{\Delta_{\min}^2}
                            + \frac{K n^{1/3}}{e^2}
                            +
                           \frac{8 N^3 K \ln{(\frac{n}{K} N^2)}}{\Delta_{\min}^2} {n^{\frac{1}{3}}}
                                        + \frac{KN}{(1-1/e)\Delta_{\min}^2}
                           \biggl)
        \biggl\}
    \end{eqnarray}
\end{theorem}

\section{$\beta$-Approximation Distribution-free Learning Policy}
\label{sec:beta}
As many problems in Expression (\ref{e-max}) are NP-hard  due to complex
constraints imposed to the maximum problem, it is necessary to analyze
the regret bound for the case of solving (\ref{e-max}) with
approximation algorithms. Without loss of generality, given an
algorithm  with
approximation factor  $\beta$ to solve problem in (\ref{e-max}), the
learning policy DFL becomes $\beta$-approximation policy DFL.
We consider an upper bound of all
$\beta$-approximation DFL policies. In that case we may have $\cup
F_{\beta} \subseteq F$. Thus we drop superscript $\beta$ for
$F_{\beta}$, $\mathbf{s}_{\beta,x}$, $T_{\beta,x}$   according to the context.

\begin{lemma}
    \label{lemma-beta-free}
    The $\beta$-approximation DFL policy satisfies
    \setlength\arraycolsep{2pt}
    {\small{
    \begin{eqnarray}
     \sup \mathfrak{R}_{\beta}(n)
                    &\leq&
    \label{max}
      N K/{\beta}
          +
            \sqrt{e K}n^{\frac{2}{3}}
          + \frac{16N^3 n^{\frac{3}{4} }}{ \beta}
          + \left(
                  1
                + \frac{4 \sqrt{K} N^2}{  \beta^2}
                +  \frac{K}{e^2N}
            \right)\frac{N^2 K}{\beta}n^{\frac{5}{6}}
    \end{eqnarray}
    }}
      without dependency on $\Delta_{\beta,\min}$. The supremum is taken over all $X$-tuple of probability distributions on $[0,1]$.
\end{lemma}
\emph{Proof:} See Appendix.

For the sake of achieving a tighter bound in Theorem \ref{the:beta}, we also provide the following regret bound with dependency on $\Delta_{\beta,\min}$.

\begin{lemma}
\label{lemma-beta-depend}
   The $\beta$-approximation DFL policy satisfies
   \setlength\arraycolsep{1.5pt}
   {\small{
    \begin{eqnarray}
        \mathfrak{R}_{\beta}(n)
                     &\leq&
                      \frac{e^3 K^3}{\Delta_{\beta,\min}^5}
    + \frac{NK}{\beta}
    \biggl(
         1 + \frac{16N^2  \ln{(\frac{n^{2/3}}{K} N^2)}}{\Delta_{\beta,\min}^2}
                            + \frac{K n^{1/3}}{e^2}
    +
        {8 N^3 K  n^{\frac{1}{3}}} \frac{\ln{(\frac{n}{k} N^2)}}{ \beta^2 \Delta_{\beta,\min}^2}
        +   \frac{N K}{(1-1/e)\Delta_{\beta,\min}^2}
    \biggl).
    \end{eqnarray}
    }}
\end{lemma}
\emph{Proof:}
    See Appendix.

Lemma \ref{lemma-beta-free} and \ref{lemma-beta-depend} together imply the following theorem under $\beta$-approximation  DFL.
\begin{theorem}
\label{the:beta}
The regret of $\beta$-approximation  DFL is bounded by
    \begin{eqnarray}
   \mathfrak{R}_{\beta}(n)
                     \leq    \min \biggl\{
                N K/{\beta}
          +
            \sqrt{e K}n^{\frac{2}{3}}
          + \frac{16N^3 n^{\frac{3}{4} }}{\beta}
          + \left(
                  1
                + \frac{4 \sqrt{K} N^2}{  \beta^2}
                +  \frac{K}{e^2N}
            \right)\frac{N^2 K}{\beta}n^{\frac{5}{6}}
        ,  \nonumber \\
         \frac{e^3 K^3}{\Delta_{\beta,\min}^5}
    + \frac{NK}{\beta}
    \biggl(
         1 + \frac{16N^2  \ln{(\frac{n^{2/3}}{K} N^2)}}{\Delta_{\beta,\min}^2}
                            + \frac{K n^{1/3}}{e^2}
    +
        {8 N^3 K  n^{\frac{1}{3}}} \frac{\ln{(\frac{n}{k} N^2)}}{ \beta^2 \Delta_{\beta,\min}^2}
        +   \frac{N K}{(1-1/e)\Delta_{\beta,\min}^2}
    \biggl)
        \biggl\}
    \end{eqnarray}
\end{theorem}

Based on Lemma~\ref{lemma-beta-free}, one can design efficient algorithms  on strategy decision even though the number
of strategies may grow exponentially.
In unknown stochastic environment, many network optimization problems
can be formulated as a  linearly combinatorial MAB problem with a
maximum objective function, e.g, the shortest path problem,  matching
problem, maximum weighted independent set of vertices problem and
other practical problems in wireless communication. For these problems which do not admit optimal solutions in polynomial time, our results provide an alternative approximation learning
methods with bounded $\beta$-regret.

\section{A Revisit to Motivation Examples}
\label{sec:application}
    In this section, we show how to leverage our proposed learning policy to tackle previous motivation applications listed in Section \ref{sec:motivation}.

%

\subsection{Online Ad placement}


   In online ad placement, the ad agent selects $N$ categories of ads from $K$ to display to targeted users, each associated with a bid $b_i$. The user's interest on each category of ads is unknown, described as a random process with average click throughput rate $p_i$. Every time the user visits the website, the agent adaptively selects a set of at most $N$ ads with distinct categories to maximize the longtime click through rate. The sum of bids on selected ads must be above a threshold $h$ to ensure the agent's profit, i.e.,
   \begin{equation}
        \sum_{s_{x,i} \in \mathbf{s}_x} b_i > h.
   \end{equation}
    In this application scenario, user's click behavior on each class of ads $\xi_i(t)=\{0,1\}$ is observed after display, and required to learn by the ad agent. The learning approach in Algorithm~\ref{alg-ad} can be applied.

    \begin{algorithm}[ht]
        \caption{Online Ads placement}
        \label{alg-ad}
            \begin{algorithmic}[1]
                \STATE For each time slot $t=0,1,\dots,n$ \\*
                    Select a set of ads $\mathbf{s}_x$ by maximizing
                                          \begin{equation}
                            \max_{\mathbf{s}_x\in F} \sum_{s_{x,i} \in \mathbf{s}_x}
                                \biggl( \tilde{\mu}_{s_{x,i}}(t)
                                     +\sqrt{
                                        \frac{ \max{(\ln{\frac{t^{2/3}}{K m_{s_{x,i}}} }},0)} {m_{s_{x,i}}}
                                      }
                                \biggl)
                        \end{equation}

            \end{algorithmic}
            \end{algorithm}

\subsection{Stochastic shortest path problem}

For this problem, we can look upon delay of each edge as a bandit.
The shortest path problem involves a minimum problem that is the opposite of maximum problems in our paper.
Thus we can transform it into a maximum problem by replacing the loss of delay with a gain
 that is defined as the difference between the maximum delay and observed delay.
Let delay of each edge be an i.i.d stochastic process $\xi_{k}(t)$ over time with mean $\mu_{k}$.
For simplicity, we assume $\xi_{k}(t)$ is normalized to $[0,1]$.
Define $\vartheta_{k}(t) = 1- \xi_{k}(t)$ with mean $1- \mu_{k}$.
We suppose each source-destination path $\mathbf{p}_{x} \in F$
   consists of a sequence of edges $\{p_{x,i}| p_{x,i} \leq |E|\}$
   where $p_{x,i}$ is index of edges.
Thus the solution to shortest path problem solves the following maximum problem actually,
  \begin{eqnarray}
    \max_{ \mathbf{p}_x \in F} \sum_{p_{x,i} \in \mathbf{p}_x} (1-\mu_{p_{x,i}})  \nonumber\\
    s.t. \ F \mbox{ is a set of all source-destination paths.}
  \end{eqnarray}

Taking $\vartheta_{k}(t)$ as unknown random variables,
  and $\mathbf{p}_x$ as strategies,
  we instantly get the maximum reward version of combinatorial multi-armed bandit formulation.
The modified DFL policy for the shortest path problem is shown in Algorithm~\ref{alg-sp}.
For the shortest path problem in (\ref{sp}) where estimation of delay on each edge is
  $\tilde{\mu}_{p_{x,i}}(t)+\sqrt{\frac{ \max{(\ln{\frac{t^{2/3}}{K m_{p_{x,i}}} }},0)} {m_{p_{x,i}}}}$,
  there exist efficient implementations of these classical solutions
  (i.e., Dijkstra's algorithm\citep{mohring2005partitioning}\citep{crauser1998parallelization}
   and Bellman-Ford algorithm\citep{goldberg1993heuristic}).
 \begin{algorithm}[ht]
        \caption{Learning policy for shortest path problem}
        \label{alg-sp}
            \begin{algorithmic}[1]
                \STATE For each time slot $t=0,1,\dots,n$ \\*
                    Select a path $\mathbf{p}_x$ by minimizing
                                        \begin{equation}
                            \label{sp}
                            \min_{\mathbf{p}_x\in F} \sum_{p_{x,i} \in \mathbf{p}_x}
                                \biggl( \tilde{\mu}_{p_{x,i}}(t)
                                     +\sqrt{
                                        \frac{ \max{(\ln{\frac{t^{2/3}}{K m_{p_{x,i}}} }},0)} {m_{p_{x,i}}}
                                      }
                                \biggl)
                        \end{equation}

            \end{algorithmic}
            \end{algorithm}


\subsection{Dynamic channel accessing in multi-hop cognitive radio networks}
\label{app-spa}


\begin{figure}[t]
\centering
    \includegraphics[width=6cm]{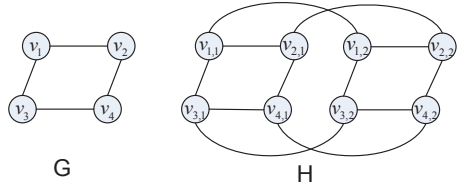}
\caption{Original conflict graph $G$ to extended conflict graph $H$.}
\label{mab-gtoh}
\vspace{-20pt}
\end{figure}

We then show how the dynamic channel accessing  problem can be formulated into a networked multi-armed bandit problem.
We remodel the network conflict graph $G$ as  an extended conflict graph $H$,
  and show that the problem can be reformulated
  as the maximum weighted independent set of vertexes in extended conflict graph $H$.
Define virtual nodes $v_{i,j}$, $j=1,\dots,M$ for each user $v_i$,
 and connect $v_{i,j}$ with $v_{i,k}(j\neq k)$ for all $j,k$.
We also connect $v_{i,j}$ with $v_{p,j}$
 if $v_i$ and $v_p$ has an edge in original network $G$.
Then we get a new graph $H$ with $MN$ nodes.
We give an illustration of this procedure in Fig.~\ref{mab-gtoh},
  where the original conflict graph $G$ has $M=2$ available channels for each of $N=4$ user.
The feasible strategy set $F$ consists of all maximal independent set (\textbf{MIS}) of nodes in $H$.
Here note that the cardinality of MIS is less than $N$
  if the chromatic number of $G$ is greater than $M$, and is $N$ otherwise.
Let $\xi_{i,j}(t)$ be weight of virtual node $v_{i,j}$.
If the mean of $\xi_{i,j}(t)$ is known,
  the optimum strategy is to find a maximum weighted independent set of nodes among $K=MN$ nodes of $H$ as choices selected by users in $G$, i.e,
\begin{eqnarray}
      &&         \max_{\mathbf{s}_x\in F} \sum_{i=1}^{N} \mu_{i,s_{x,i}} \nonumber\\
 &&\label{mwis}      s.t.~  \mathbf{s}_x \mbox{ is an independent set of vertexes in } H,
\end{eqnarray}
where $s_{x,i}$ is the index of channel selected by user $v_i$ in strategy $\mathbf{s}_x$.

 \begin{algorithm}[!htp]
        \caption{Dynamic channel accessing in multi-hop cognitive radio networks}
        \label{alg-dcs}
            \begin{algorithmic}[1]
                \STATE For each time slot $t=0,1,\dots,n$ \\*
                    Select a strategy $\mathbf{s}_x$  by minimizing
                                        \begin{equation}
                            \label{estimated-mwis}
                             \sum_{s_{x,i} \in \mathbf{s}_x}
                                \biggl( \tilde{\mu}_{s_{x,i}}(t)
                                     +\sqrt{
                                        \frac{ \max{(\ln{\frac{t^{2/3}}{K m_{s_{x,i}}} }},0)} {m_{s_{x,i}}}
                                      }
                                \biggl)
                        \end{equation}

            \end{algorithmic}
            \end{algorithm}

Similarly, the dynamic channel accessing policy in Algorithm~\ref{alg-dcs} needs to find a strategy that has maximum estimated weight at each time slot, i.e., solving the problem of (\ref{estimated-mwis}),
where
    \begin{equation*}
    \tilde{\mu}_{s_{x,i}}(t)
                                     +\sqrt{
                                        \frac{ \max{(\ln{\frac{t^{2/3}}{K m_{s_{x,i}}} }},0)} {m_{s_{x,i}}}
                                      }
    \end{equation*}
    is estimated weight of virtual node $v_{\mathbf{s}_{x,i}}$.
As the involved MWIS problem is NP-hard, we can not directly use the DFL policy to solve (\ref{estimated-mwis}).
We then turn to $\beta$-approximation DFL policy to solve (\ref{estimated-mwis}) with low complexity approximation algorithms for MWIS.
For MWIS problem, there exist some simple PTAS that can be implemented in a distributed manner,
 such as robust PTAS in \citep{nieberg2005robust} and shifting approach in \citep{mwis2005}.

Herein the above applications give basic  frameworks on bandits formulation of these problems,
 practical considerations may generate  more complicated constraints on feasible sets $F$,
 which lead to even harder NP problems that have no existing efficient solutions.
Additionally, many more details and implementation issues  need to be addressed
 when applying our proposed policies to specific applications.
For instance, in the application of dynamic channel accessing,
 it would be necessary to design a local or distributed implementation of our policy,
 involving consideration on low cost on strategy decision, as well as message collection and broadcast.
These issues are not trivial, but of significance when putting our theoretical results into practice.
  It demands a careful tradeoff among theoretical guarantee, implementation manners, storage, computation and extra communication complexity
    as well as their potential impact on the actually achievable performance.
Hence, combination of practical implementation with our proposed learning policy in specific domain especially demands more elegant design,
 which is also an interesting work. 

\section{Simulation}
\label{sec:simulation}
In addition to obtaining the regret bounds of our learning policy, we are also interested in understanding its performance in practise. In this section, we present some simulation results by applying DFL to ad placement problem and  dynamic channel accessing problem as described in Section~\ref{sec:application}.

  \subsection{Online ad placement problem}
We consider a website with $5$ ad placements targeted at  users.
We assume that there are $10$ categories of advertisements.
The bids and a specific user's interests (denoted by click-through-rates that are unknown)  for each category are shown in Table~\ref{tbl-ad}.
The threshold is set as $3000$, so the static optimum is $3.8414$ with  the set of ad categories $\{1,2,4,5,9 \}$, if the ad agency knows user's interest.

\begin{figure}[t]
    \centering
            \includegraphics[ width=3in]{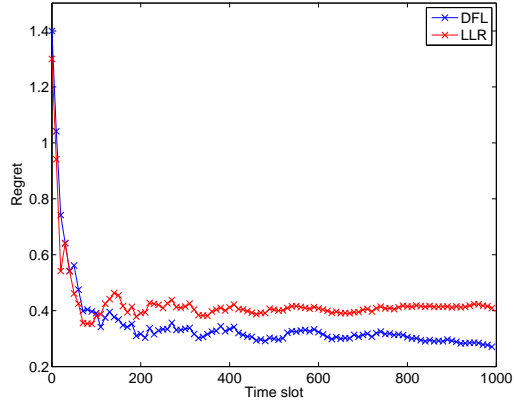}
   \caption{Regret in ad placement: comparison with LLR learning policy  }
\label{regret-ad}
\end{figure}

We compare DFL
to one state-of-the-art approach LLR for the time averaged-regret.
 Fig.~\ref{regret-ad} shows the comparison results.
We find that DFL achieves significant performance gains over LLR in terms of lower regret.

\begin{table}[t]\setlength{\tabcolsep}{3pt}
\begin{center}
\caption{Bids and user's interests of each category of ads}
\label{tbl-ad}
\begin{tabular}{c|c| c}
\hline
   Category & Bid & Avg Click Through Rate\\
    \hline
            0&0.4506&640.9853 \\
            1&0.7279&173.41842\\
            2&0.8377&924.09434\\
            3&0.1662&601.3466\\
            4&0.8055&705.72878\\
            5&0.7732&759.04837\\
            6&0.2179&302.2392\\
            7&0.2688&809.4084\\
            8&0.3722&421.9816\\
            9&0.6971&771.5156\\
    \hline
\end{tabular}
\end{center}
\end{table}

 \subsection{Dynamic channel accessing problem}
In this round of experiment, we evaluate the performance of our policy in the context of dynamic channel accessing problem. We consider a small network with $5$ users, each of which has $5$ available channels.
The conflict relationship is below,
\begin{displaymath}
\label{conflicts}
\left(\begin{array}{c c c c c}
    1 & 1 & 1 & 1 & 0           \\
    1 & 1 & 1 & 0 & 1             \\
    1 & 1 & 1 & 1 & 0             \\
    1 & 0 & 1 & 1 & 0             \\
    0 & 1 & 0 & 0 & 1
  \end{array}\right)
\end{displaymath}
 where an element $e_{i,j}=1$ denotes conflict and $e_{i,j}=0$ denotes independency between users $v_i$ and $v_j$.
The average data rate on the $5$ channels of each user is shown in the following matrix,
\begin{displaymath}
\label{weightsOfChannels}
\left(\begin{array}{c c c c c}
    631.98 & 369.81 & 128.43 & 191.70 & 155.64           \\
    432.00 & 53.93 & 598.08 & 30.93 & 551.52             \\
    199.55 & 26.00 & 1175.17 &  524.34 & 147.69             \\
    127.38 & 53.73 &  68.34 & 937.44 & 117.62             \\
    311.04 & 101.28 & 171.95 &  436.45 &  62.19
  \end{array}\right)
\end{displaymath}
where
each row $i$ denotes data rates of user $v_i$, $i=1,2,3,4,5$.
The optimal static throughput of this network, i.e.,
 the maximum possible weight of  ISLs in the corresponding extended conflict graph  is $3732.56$.

\begin{figure}[t]
    \centering
            \includegraphics[ width=3in]{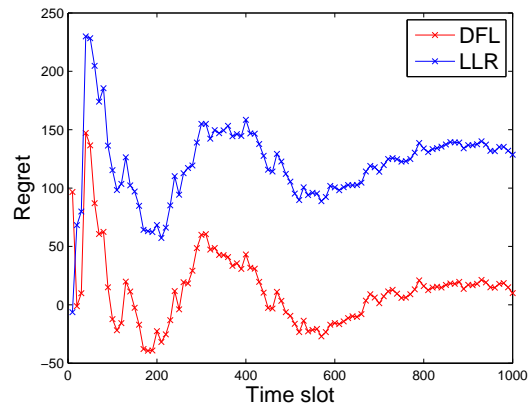}
   \caption{Regret in dynamic channel accessing: comparison with LLR learning policy  }
\label{regret-opt}
\end{figure}

\begin{figure}[t]
    \centering
        \includegraphics[ width=3in]{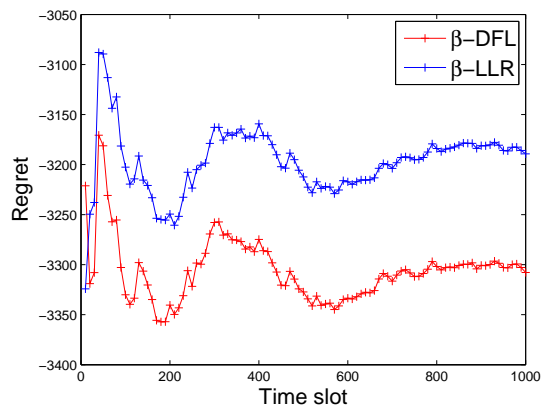}
   \caption{$\beta$-regret in dynamic channel accessing: comparison with LLR learning policy  }
   \label{regret-beta}
\end{figure}

Fig.~\ref{regret-opt} and Fig.~\ref{regret-beta} plot comparison of the time-averaged regret/$\beta$-regret by our proposed
 DFL policy and LLR policy. Fig.~\ref{regret-opt} shows that DFL policy requires much less time on learning for better  strategies, thus produces much smaller regret.
The time-averaged regret by DFL policy converges to $0$ around time slot $400$, while regret by LLR
 policy is more than $100$.
$\beta$-regret in Fig.~\ref{regret-beta} shows negative value, which indicates that the achievable throughput
by the two learning algorithms is better than $1/\beta$ of the optimal throughput when utilizing $\beta$-approximation algorithms to solve the NP-hard MWIS problem.

\section{Conclusion and Discussion}
\label{sec:conclusion}
In this paper we propose a distribution-free policy for arbitrary linearly combinatorial multi-armed bandits with general constraints that may cause  exponential number of strategies. We  have taken care of efficiency issues on storage, computation and practical applications. We expect that our works would broaden applications of multi-armed bandits in practice.

We admit that the theoretical regret bound is kind of loose compared to some existing results.
It is interesting that  the simulation results actually show obvious advance on  LLR policy  \citep{gai2012mab},
though the theoretical result bound of LLR achieving a regret logarithmal with time.
Therefore, better results are probably available through other techniques.
The limitation of our theoretical analysis lies in the peeling argument that we adopt to derive for distribution-free bound.
Current form of the function $f(x)$ makes it impossible to get a regret bound with a smaller order of time.
We conjecture that an upper bound with  $O(\sqrt{n})$ may be available if we design better functions in the peeling argument and use more strict conditions in probabilistic analysis when counting the number of times that non-optimal or non-$\beta$-approximation strategies have been played.
We leave this challenging question as a future work.

In our paper we have actually studied a simpler bandit model of stochastic rewards, compared to adversary or Markovian bandits. The assumption on \emph{i.i.d} stochastic process has mitigated difficulties on concentration analysis through Hoeffding's results. The problem becomes more challenging in the adversary case where we can not use these tools. For instance, many results with tight regret bounds of $\sqrt{nK}$ have been gained for linear combination of bandits in the adversary case in literature, but not yet computation and storage efficient.
We expect to tackle this challenge in future works.

We also note that many works as well as ours have studied \emph{weak regret} that is compared to a static optimal policy.
It would be interesting to analyze models using \emph{strong regret} that is compared to a dynamic optimal policy. In this case, one has to track the best dynamic policy through estimating the random process and computing the approximate optimal policy, instead of only estimating sample mean in static case. Similar with the case of weak regret, there would be challenges on reduction of time and space complexity, as well as distributed implementation issues among multi-users.


\bibliographystyle{ormsv080}
\bibliography{amab}

\section{Appendix}

\subsection{Proof of Lemma \ref{lemma-opt-free}}

To prove Lemma \ref{lemma-opt-free}, we need to use Chernoff-Hoeffding bound and the maximal inequality by Hoeffding \citep{hoeffding1963probability}.

\begin{lemma} {(Chernoff-Hoeffding Bound \citep{hoeffding1963probability})}
\label{lemma-chernoff}
    $\xi_1,\dots,\xi_n$ are random variables within range $[0,1]$, and $E[\xi_t|\xi_1,...,\xi_{t-1}]=\mu,\forall 1 \leq t \leq n$. Let $S_n =\sum \xi_i$, then for all $a>0$
    \begin{eqnarray}
        \mathbf{P}(S_n \geq n \mu + a) \leq \exp{(-2 a^2 /n)},    \nonumber \\
        \mathbf{P}(S_n \leq n \mu - a) \leq  \exp{(-2 a^2 /n)}.
    \end{eqnarray}
\end{lemma}

\begin{lemma}{(Maximal inequality)}\citep{hoeffding1963probability}
\label{lemma-maximal}
    $\xi_1,\dots,\xi_n$ are i.i.d random variables with expect $\mu$, then for any $y>0$ and $n>0$,
    \begin{equation}
        \mathbf{P}  \biggl(
                    \exists \tau \in {1,\dots,n}, \sum_{t=1}^{\tau} (\mu-\xi_t) >y  \biggl)< \exp(-\frac{2y^2}{n})
            .
    \end{equation}
\end{lemma}

Recall that we have assumed $\lambda_1 \geq \dots \geq \lambda_X$. As strategy $\mathbf{s}_1$ is the optimal strategy, we have $\Delta_x=\lambda_1-\lambda_x$, and let $Z_x = \lambda_1- \frac{\Delta_x}{2}$. We further define
$W_1 = \min_{1\leq t \leq n} W_1(t).$
We may assume the first time slot $z= \argmin_{1\leq t \leq n} W_1(t)$.

\textbf{1. Rewrite regret in terms of arms}

Separating the strategies in two sets by $\Delta_{x_0}$ of some strategy $\textbf{s}_{x_0}$(we will define $x_0$ later in the proof), we have
\begin{eqnarray}
    \mathfrak{R}(n) &=&    \sum_{x=1}^{x_0} \Delta_x E[T_x(n)] + \sum_{x=x_0+1}^{X} \Delta_x E[T_x(n)] \nonumber\\
   \label{rn}             &\leq&  \Delta_{x_0}n + \sum_{x=x_0+1}^{X} \Delta_x E[T_x(n)] .
\end{eqnarray}
We then analyze the second term of (\ref{rn}). As there may be exponentially large number of strategies, counting $T_x(n)$ of each strategy by the traditional UCB based analysis yields regret growing linearly with the number of strategies. Note that each strategy consists of $N$ arms at most, we can rewrite the regret in terms of arms instead of strategies. We then introduce a set of counters $\{\widetilde{T}_k(n)|k=1,\dots,K\}$. At each time slot, either 1) a strategy with $\Delta_x \leq \Delta_{x_0}$ or 2) a strategy with $\Delta_x > \Delta_{x_0}$ is played. In the first case, no $\widetilde{T}_k(n)$ will get updated. In the second case, we increase $\widetilde{T}_k(n)$ by $1$ for any arm $k= \argmin_{s_{x,j}\in \mathbf{s}_x}\{ m_{s_{x,j}}\}$. Thus whenever a strategy with $\Delta_x > \Delta_{x_0}$ is chosen, exactly one element in $\{\widetilde{T}_k(n)\}$ is increased by $1$. This implies that the total number that strategies of $\Delta_x > \Delta_{x_0}$ have been played is equal to sum of all counters in
$\{\widetilde{T}_k(n)\}$, i.e., $\sum_{x=x_0+1}^{X} E[T_x(n)] = \sum_{k=1}^{K} \widetilde{T}_k(n)$. Thus, we can rewrite the second term of (\ref{rn}) as
\begin{eqnarray}
       \sum_{x=x_0+1}^{X} \Delta_x E[T_x(n)]
   \leq \Delta_{X} \sum_{x=x_0+1}^{X}E[T_x(n)]
\label{txtk}   &\leq& \Delta_{X} \sum_{k=1}^{K} E[ \widetilde{T}_k(n)]. \nonumber\\
\end{eqnarray}
Let ${I}_k(t)$ be the indicator function that equals $1$ if $\widetilde{T}_k(n)$ is updated at time slot $t$. Define the indicator function $\mathbf{1}\{y\} =1$ if the event $y$ happens and $0$ otherwise. When  ${I}_k(t)=1$, a strategy $\mathbf{s}_x$ with $x>x_0$ has been played for which $m_k = \min \{m_{s_{x,j}}: \forall s_{x,j} \in \mathbf{s}_x\}$.
Then
\begin{eqnarray}
   \widetilde{T}_k(n) &=& \sum_{t=1}^{n} \mathbf{1} \{{I}_k(t)=1\} \\
                       &\leq& \sum_{t=1}^{n} \mathbf{1} \{W_1(t) \leq W_x(t)\} \\
                       &\leq& \sum_{t=1}^{n} \mathbf{1} \{W_1 \leq W_x(t)\} \\
     \label{tk1}                  &\leq& \sum_{t=1}^{n} \mathbf{1} \{W_1 \leq W_x(t), W_1 \geq  Z_x \} \\
     \label{tk2}                  &~&       + \sum_{t=1}^{n} \mathbf{1} \{W_1 \leq W_x(t), W_1 < Z_x \} \\
     \label{tk}                 &=&  \widetilde{T}^{1}_k(n) + \widetilde{T}^{2}_k(n).
\end{eqnarray}
We use $\widetilde{T}^{1}_k(n)$ and $\widetilde{T}^{2}_k(n)$ to respectively denote Equation (\ref{tk1}) and (\ref{tk2}) for short. Next we show that both of the terms are bounded.

2. \textbf{Bounding $\widetilde{T}^{1}_k(n)$}

Here we note the event  $\{W_1 \geq Z_x\}$ and $\{W_x(t) > W_1 \}$ implies event $\{W_x(t) > Z_x\}$. Let $\ln_+(y)= \max(\ln(y),0)$.
For any positive integer $l_0$, we then have,
\begin{eqnarray}
    \widetilde{T}^{1}_k(n) &\leq& \sum_{t=1}^{n} \mathbf{1}\{W_x(t) \geq Z_x\}  \\
                           &\leq& l_0 + \sum_{t=l_0}^{n} \mathbf{1}\{W_x(t) \geq Z_x, \widetilde{T}^{1}_k(t)> l_0\} \\
                           &=& l_0 + \sum_{t=l_0}^{n} \mathbf{P} \{W_x(t) \geq Z_x, \widetilde{T}^{1}_k(t)> l_0\} \\
                           &=& l_0 + \sum_{t=l_0}^{n}
     \label{tk1p3}                     \mathbf{P} \biggl\{ \sum_{s_{x,j} \in \mathbf{s}_x} \biggl(\tilde{\mu}_{s_{x,j}} + \sqrt{\frac{\ln_+( \frac{t^{2/3}}{K m_{s_{x,j}}})} {l_0}} \biggl) \nonumber\\
                           &~&                \geq \sum_{s_{x,j} \in \mathbf{s}_x} {\mu}_{s_{x,j}}+ \frac{\Delta_x}{2}, \widetilde{T}^{1}_k(t)> l_0
                                          \biggl\}.
\end{eqnarray}
The event $\biggl\{ \sum_{s_{x,j} \in  \mathbf{s}_x} \biggl( \tilde{\mu}_{s_{x,j}} + \sqrt{\frac{\ln_+(t^{2/3}/Km_{s_{x,j}})}{m_{s_{x,j}}}} \biggl) \geq \sum_{s_{x,j} \in \mathbf{s}_x} {\mu}_{s_{x,j}}+ \frac{\Delta_x}{2} \biggl\} $
indicates that
\begin{equation}
    \exists s_{x,j} \in \mathbf{s}_x, \tilde{\mu}_{s_{x,j}} + \sqrt{\frac{\ln_+(t^{2/3}/K m_{s_{x,j}})}{m_{s_{x,j}}}} \geq  {\mu}_{s_{x,j}}+ \frac{\Delta_x}{2N}.
\end{equation}
Using union bound one directly obtains:
\begin{eqnarray}
    \widetilde{T}^{1}_k(n) &\leq& l_0 +  \sum_{t=l_0}^{n} \sum_{s_{x,j} \in \mathbf{s}_x}
                                \mathbf{P} \biggl\{ \tilde{\mu}_{s_{x,j}} + \sqrt{\frac{\ln_+(t^{2/3}/K m_{s_{x,j}})}{m_{s_{x,j}}}}
                                          \geq  {\mu}_{s_{x,j}}+ \frac{\Delta_x}{2N} \biggl\} \\
                        \label{tk1p1}     &\leq& l_0 +  \sum_{t=l_0}^{n} \sum_{s_{x,j} \in \mathbf{s}_x}   \mathbf{P} \biggl\{ \tilde{\mu}_{s_{x,j}} - {\mu}_{s_{x,j}}
                           \geq  \frac{\Delta_x}{2N}  - \sqrt{\frac{\ln_+(t^{2/3}/Km_{s_{x,j}})}{m_{s_{x,j}}}} \biggl\}.
\end{eqnarray}

Let $l_0=  16N^2 \lceil \ln(\frac{n^{3/4}}{K} \Delta_{x}^2) / \Delta_{x}^2) \rceil $, where the notation $\lceil y \rceil$ represents the smallest integer that is larger than $y$. We further set $ \delta_0 = e^{1/2} \sqrt{K/n^{2/3}}$ and set $x_0$ such that $\Delta{x_0} \leq \delta_0 < \Delta_{x_0 +1}$.
As $m_{s_{x,j}} \geq l_0$, we have
\begin{eqnarray}
    & &\ln_+ \biggl(\frac{t^{2/3}}{K m_{s_{x,j}}} \biggl)
                                \leq \ln_+ \biggl(\frac{n^{2/3}}{K m_{s_{x,j}}} \biggl)
                                \leq \ln_+ (n^{2/3}/K l_0)                              \nonumber \\
                                 & \leq & \ln_+ ( \frac{n^{2/3}}{K} \times \frac{\Delta_x^2}{16N^2} )
                                   \leq\frac{l_0 \Delta_x^2}{16N^2} \leq \frac{m_{s_{x,j}} \Delta_x^2}{16N^2}.
\end{eqnarray}
Hence we have,
\begin{equation}
    \frac{\Delta_x}{2N} - \sqrt{\frac{\ln_+(t^{2/3}/Km_{s_{x,j}})}{m_{s_{x,j}}}} \geq \frac{\Delta_x}{2N} - \frac{\Delta_x}{\sqrt{16N^2}} = c \Delta_x
\end{equation}
with $c=\frac{1}{2N}-\frac{1}{\sqrt{16N^2}} = \frac{1}{4N}$.

Therefor, using Hoeffding's inequality and Equation (\ref{tk1p1}), and then plugging into the value of $l_0$, we get
\begin{eqnarray}
\widetilde{T}^{1}_k(n) &\leq & l_0 + \sum_{t=l_0}^{n} \sum_{s_{x,j} \in \mathbf{s}_x}   \mathbf{P}
                                \biggl\{ \tilde{\mu}_{s_{x,j}} - {\mu}_{s_{x,j}}  \geq c \Delta_x \biggl\} \nonumber\\
                        &\leq& l_0 + \sum_{t=l_0}^{n} \sum_{s_{x,j} \in \mathbf{s}_x}  \exp(-2m_{s_{x,j}}(c\Delta_x)^2) \nonumber\\
                        &\leq& l_0 + K \cdot n \cdot  \exp(-2l_0(c\Delta_x)^2) \nonumber \\
   \label{tk1p2}        &=& 1+ 16N^2 \frac{\ln(\frac{n^{3/4}}{K} \Delta_{x}^2)}{ \Delta_{x}^2}+  K \cdot n \cdot \exp(-2\ln(n^{\frac{1}{12}}e)).
\end{eqnarray}

 As $\delta_0 = e^{1/2} \sqrt{K/n^{\frac{2}{3}}}$,
 and
  $ x\ln(\frac{n^{3/4}}{K} \Delta_{x}^2)/\Delta_{x}^2$
achieves the maximum value of $\exp(-K/n^{\frac{3}{4}})$ when
 $\Delta_x = \sqrt{ K e^{K/n^{\frac{3}{4}}}/n^{\frac{3}{4}}  }$, the second term in (\ref{tk1p2}) is bounded by
\begin{eqnarray}
    \frac{16N^2}{\exp(K/n^{3/4})} < \frac{16N^2}{1+K/n^{3/4}} < \frac{16N^2 n^{3/4}}{K}
\end{eqnarray}

The last term of (\ref{tk1p2}) is bounded by
\[K \cdot n \cdot \exp(-2\ln(n^{\frac{1}{12}}e)) \leq \frac{K}{e^2}\cdot n^{\frac{5}{6}}\]
Finally we get
{\small{
\begin{eqnarray}
    \widetilde{T}^{1}_k(n) \leq 1+  \frac{16N^2 n^{3/4} }{K} +   \frac{K}{e^2}\cdot n^{\frac{5}{6}}.
\end{eqnarray}
}}
3. \textbf{Bounding $\widetilde{T}^{2}_k(n)$}
{\small{
\begin{eqnarray}
    \widetilde{T}^{2}_k(n)  &=&     \sum_{t=1}^{n} \mathbf{1}\{ W_1 \leq W_x(t), W_1 <  Z_x \}                      \nonumber \\
                            &\leq&  \sum_{t=1}^{n} \mathbf{P} \{ W_1 < Z_x\}
  \label{tk2n}                          \leq  n\mathbf{P} \{ W_1 < Z_x\}.
\end{eqnarray}
}}
Remember that at time slot $z$, we have $W_1 =\min{W_1(t)}$. For  the probability $\{W_1 < Z_x\}$ of fixed $x$, we have
\begin{eqnarray}
                    &~&     \mathbf{P}  \{W_1 < \lambda_1 - \frac{\Delta_x}{2}\}                                                                       \\
      &=&     \mathbf{P}  \biggl\{\sum_{s_{1,j} \in \mathbf{s}_1, j=1}^{N} w_{s_{1,j}} (z)<  \lambda_1 - \frac{\Delta_x}{2} \biggl\}           \\
 \label{Tk2p1} &\leq&  \sum_{s_{1,j}\in \mathbf{s}_1} \mathbf{P} \biggl\{w_{s_{1,j}}(z) <  \mu_{s_{1,j}} - \frac{\Delta_x}{2N}   \biggl\}.
    \end{eqnarray}
We define function $f(u)= e \ln(\sqrt{\frac{n^{1/3}}{K}} u)/u^3$ for $u \in [\delta_0,N]$. Then we have,
\begin{eqnarray}
   &~&  \mathbf{P}  \biggl\{w_{s_{1,j}}(z) < \mu_{s_{1,j}} - \frac{\Delta_x}{2N}    \biggl\}                           =  \mathbf{P}  \biggl\{\exists 1 \leq l \leq n: \sum_{\tau=1}^{l} \biggl(\xi_{s_{1,j}}(\tau)
              + \sqrt{\frac{\ln_+(\frac{\tau^{2/3}}{K l})} {l} } \biggl)                   < l\mu_{s_{1,j}} - \frac{l\Delta_x}{2N}  \biggl\}                                                    \nonumber \\
   &\leq& \mathbf{P}    \biggl\{
                       \exists 1 \leq l \leq n: \sum_{\tau=1}^{l} (\mu_{s_{1,j}}- \xi_{s_{1,j}}(\tau))    >   \sqrt{l\ln_+(\frac{\tau^{2/3}}{K l})} +  \frac{l\Delta_x}{2N}
                \biggl\}                                                                                        \nonumber \\
   &\leq& \mathbf{P}    \biggl\{
                        \exists 1 \leq l \leq f(\Delta_x):                                                             \sum_{\tau=1}^{l} (\mu_{s_{1,j}}- \xi_{s_{1,j}}(\tau))
                        >\sqrt{l\ln_+(\frac{\tau^{2/3}}{K l})}
                \biggl\}                                                                                         \nonumber \\
   &~& +  \mathbf{P}    \biggl\{
                        \exists f(\Delta_x) < l \leq n:                                                             
                            \sum_{\tau=1}^{l} (\mu_{s_{1,j}}- \xi_{s_{1,j}}(\tau))
                        > \frac{l\Delta_x}{2N}
                \biggl\}.
\end{eqnarray}
For the first term we use a peeling argument with a geometric grid of the form $\frac{1}{2^{g+1}} f(\Delta_{x}) \leq l \leq \frac{1}{2^{g}} f(\Delta_{x})$:
\begin{eqnarray}
    &&    \mathbf{P}      \biggl\{
                        \exists 1 \leq l \leq f(\Delta_x): \sum_{\tau=1}^{l} (\mu_{s_{1,j}}- \xi_{s_{1,j}}(\tau))       
                        >\sqrt{l\ln_+(\frac{\tau^{2/3}}{K l})}
                \biggl\}                                                                                                   \nonumber\\
    &\leq& \sum_{g=0}^{\infty} \mathbf{P}\biggl\{
                       \exists  \frac{1}{2^{g+1}} f(\Delta_{x}) \leq l \leq \frac{1}{2^{g}} f(\Delta_{x}):               \sum_{\tau=1}^{l} (\mu_{s_{1,j}}- \xi_{s_{1,j}}(\tau))
                       >\sqrt{ \frac{f(\Delta_x)}{2^{g+1}}\ln_+(\frac{\tau^{2/3}2^g}{K f(\Delta_{x})})}
                \biggl\}                                                                                                       \nonumber\\
    &\leq&   \sum_{g=0}^{\infty} \exp   \biggl(
                                               -2\frac{ f(\Delta_x) \frac{1}{2^{g+1}} \ln_+(\frac{\tau^{2/3} 2^g}{K f(\Delta_x)})} {f(\Delta_x)\frac{1}{2^g}}
                                        \biggl)                                                                                 \nonumber\\
    &\leq&     \sum_{g=0}^{\infty} \biggl[
                                        \frac{K f(\Delta_x)}{n^{2/3}} \frac{1}{2^g}
                                \biggl]
    \leq
 \label{tk2r1}                                       \frac{2 K f(\Delta_x)}{n^{2/3}}
\end{eqnarray}
where in the second inequality we use  Lemma~\ref{lemma-maximal}.

Due to the special design of function $f(u)$, we have $f(u)$ achieves the maximum value of $\frac{n^{1/2}}{3 K^{3/2}} $ when $u=e^{1/3}\sqrt{K/n^{1/3}}$.
We then have
\begin{eqnarray}
      \frac{2K f(\Delta_x)}{n^{2/3}} \leq \frac{2}{3 \sqrt{K}} n^{-1/6}.
\end{eqnarray}
For the second term, we also use a peeling argument but with a geometric grid of the form $2^g f(\Delta_x) \leq l < 2^{g+1} f(\Delta_x)$:
\begin{eqnarray}
    &&  \mathbf{P}      \biggl\{
                        \exists f(\Delta_x) < l \leq n: \sum_{\tau=1}^{l} (\mu_{s_{1,j}}- \xi_{s_{1,j}}(\tau))
                        > \frac{l\Delta_x}{2N}
                \biggl\}                                                                                                    \nonumber\\
    &\leq&      \sum_{g=0}^{\infty} \mathbf{P}\biggl\{
                                        \exists 2^{g} f(\Delta_x) \leq l \leq 2^{g+1} f(\Delta_x):                                                             \sum_{\tau=1}^{l} (\mu_{s_{1,j}}- \xi_{s_{1,j}}(\tau))> \frac{2^{g-1}f(\Delta_x)\Delta_x}{N}
                                    \biggl\}                                                                                \nonumber\\
    &\leq&      \sum_{g=0}^{\infty} \exp    \biggl(
                                                   \frac{ -2^g f(\Delta_x) \Delta_x^2}{ 4N^2}
                                            \biggl)                                                                         \nonumber\\
    &\leq&      \sum_{g=0}^{\infty} \exp    \biggl(
                                                    -(g+1)f(\Delta_x) \Delta_x^2 / 4N^2
                                            \biggl)                                                                         \nonumber\\
  \label{tk2r2}  &=&         \frac{1}{\exp(f(\Delta_x)\Delta_x^2/4N^2)-1}.
\end{eqnarray}

We note that $f(u)u^2$ has a minimum value of $\frac{1}{\sqrt{K}} n^{1/6}$.
Thus for (\ref{tk2r2}), we further have,
\begin{eqnarray}
          \frac{1}{\exp(\frac{f(\Delta_x)\Delta_x^2}{4N^2})-1}  
    \leq  \frac{1}{\exp \bigg( \frac{  n^{1/6}}{4 \sqrt{K}N^2}   \biggl) -1} 
    \leq  4 \sqrt{K}N^2 n^{-\frac{1}{6}}.
\end{eqnarray}
Combining (\ref{Tk2p1}) and (\ref{tk2n}), we then have
\begin{equation}
    \widetilde{T}^{2}_k(n) \leq \frac{2N n^{5/6} }{3 \sqrt{K}} +  4 \sqrt{K}N^3 n^{5/6}
    \leq (1+ 4\sqrt{K}N^2)N n^{\frac{5}{6}}.
\end{equation}

\textbf{4. Results without dependency on $\Delta_{\min}$}

Summing $\widetilde{T}^{1}_k(n)$ and $\widetilde{T}^{2}_k(n)$,
we have
\begin{eqnarray}
    \widetilde{T}_k(n)  &\leq& \widetilde{T}^{1}_k(n) + \widetilde{T}^{2}_k(n)  \nonumber \\
                        &=&
                        1+  \frac{16N^2 n^{3/4} }{K} +   \frac{K}{e^2}\cdot n^{\frac{5}{6}}         +
                        (1+ 4 \sqrt{K}N^2) N n^{\frac{5}{6}}
\end{eqnarray}
and using  $\Delta_X \leq N$ and $\Delta_x \leq \delta_0$ for $x \leq x_0$, we have
\begin{eqnarray}
    \mathfrak{R}(n) &\leq& \sqrt{ K e} n^{\frac{2}{3}} + NK\biggl[
    1+  \frac{16N^2 n^{3/4} }{K} +   \frac{K}{e^2}\cdot n^{\frac{5}{6}}
    +   (1+ 4 \sqrt{K}N^2) N n^{\frac{5}{6}}
    \biggl]
    \nonumber\\
               &\leq& NK +
               \sqrt{ K e} n^{\frac{2}{3}} +
               16N^3 n^{3/4}  +
              \biggl[\frac{K}{e^2}+
               (1+ 4 \sqrt{K}N^2) N
              \biggl]NK n^{\frac{5}{6}}
\end{eqnarray}

\subsection{Proof of Lemma \ref{lemma-opt-depend}}

Recall that we have $\lambda_1 \geq \dots \geq \lambda_X$, and $Z_x= \lambda_1 - \frac{\Delta_x}{2}$.
This time we set $\Delta_{x_0} \leq \delta_0 = \sqrt{eK/n^{1/3}} < \Delta_{x_0+1}$.
Splitting strategy set $F$ into two disjoint sets again by $\Delta_{x_0}$, and plugging (\ref{txtk}) into (\ref{rn}), we begin with a weak vision of (\ref{rn}),
\begin{eqnarray}
    \mathfrak{R}(n)
    \label{rn_min}      &\leq&  \Delta_{x_0}n + \Delta_{X} \sum_{k=1}^{K} \widetilde{T}_k(n).
\end{eqnarray}

Here we have the same form of $\widetilde{T}_k(n)$ as that in (\ref{tk}), i.e.,
\begin{eqnarray}
        \widetilde{T}_k(n) &=& \widetilde{T}^{1}_k(n) + \widetilde{T}^{2}_k(n),           \\
        \widetilde{T}_k^1(n)    &\leq& \sum_{t=1}^{n} \mathbf{1} \{W_1 \leq W_x(t), W_1 \geq  Z_x \} , \\
        \widetilde{T}_k^2(n) &\leq& \sum_{t=1}^{n} \mathbf{1} \{W_1 \leq W_x(t), W_1 < Z_x \}.
\end{eqnarray}

Setting $l_0=16N^2 \frac{\ln(\frac{n^{2/3}}{K} \Delta_{x}^2)}{ \Delta_{x}^2}$, and similar with (\ref{tk1p2}), we have
\begin{eqnarray}
     \widetilde{T}^1_{k}(n) &\leq&
     1+
     16N^2 \frac{\ln(\frac{n^{2/3}}{K} \Delta_{x}^2)}{ \Delta_{x}^2}
     +
     \sum_{t=l_0}^{n} \sum_{s_{x,j} \in \mathbf{s}_x}  \exp(-2m_{s_{x,j}}(c\Delta_x)^2)
                                         \nonumber\\
      &\leq&
      1+
      16N^2 \frac{\ln(\frac{n^{2/3}}{K} N^2)}{ \Delta_{\min}^2}
      +
      K \cdot n \cdot \exp(-2\ln(n^{\frac{1}{3}}e))
                                            \nonumber\\
                            &\leq& 1 + \frac{16N^2  \ln{(\frac{n^{2/3}}{K} N}^2)}{\Delta_{\min}^2}
                            + \frac{K n^{1/3}}{e^2}
\label{tk1min}                             
\end{eqnarray}
where we use  $\Delta_{\min} \leq \Delta_x \leq N$ in the last term.

As to $\widetilde{T}^2_{k}(n) = n \mathbf{P} \{W_1 < Z_x\}$, according to (\ref{Tk2p1}),
\begin{eqnarray}
    \mathbf{P}(\{W_1 < Z_x\}) &\leq&
     \sum_{s_{1,j}\in \mathbf{s}_1} \mathbf{P} \biggl\{w_{s_{1,j}}(z) < \mu_{s_{1,j}} - \frac{\Delta_x}{2N} \biggl\}. 
\end{eqnarray}
Then the probability of $\{w_{s_{1,j}}(z) < \mu_{s_{1,j}} - \frac{\Delta_x}{2N}\}$ can been divide into two elements by introducing a function $f(\Delta_x) < n$. Here we again follow a similar scheme as done in proof of Lemma~\ref{lemma-opt-free}.
 We reset the function $f(\Delta_x)=4N^2 \frac{\ln{(n \Delta_x^2 /K)}}{\Delta_x^2} $, and let
 \begin{eqnarray}
    &P_1&  = \mathbf{P}   \biggl\{
                           \exists 1 \leq l \leq f(\Delta_x):                                                        
                                       \sum_{\tau=1}^{l} (\mu_{s_{1,j}}- \xi_{s_{1,j}}(\tau))
                        >\sqrt{l\ln_+(\frac{\tau^{2/3}}{K l})}
                  \biggl\},                                                                                     \nonumber  \\
    &P_2&  =  \mathbf{P}  \biggl\{
                        \exists f(\Delta_x) < l \leq n:                                                             
                               \sum_{\tau=1}^{l} (\mu_{s_{1,j}}- \xi_{s_{1,j}}(\tau))
                        > \frac{l\Delta_x}{2N}
                    \biggl\},                      \nonumber\\
    &\mathbf{P}&  \{W_1 < Z_x\} \leq  \sum_{s_{1,j}\in \mathbf{s}_1} (P_1 + P_2).  \nonumber
    \end{eqnarray}
For $P_1$ we use a peeling argument with a geometric grid of the form $\frac{1}{2^{g+1}} f(\Delta_{x}) \leq l \leq \frac{1}{2^{g}} f(\Delta_{x})$, then by using similar technique of (\ref{tk2r1}), we have
\begin{eqnarray}
     P_1 \leq \frac{2K f(\Delta_x)}{n^{2/3}} = \frac{8 N^2K}{n^{\frac{2}{3}}} \frac{\ln{(\frac{n}{K}N^2)}}{\Delta_{\min}^2} .
\end{eqnarray}
For $P_2$ we also use a peeling argument but with a geometric grid of the form $2^g f(\Delta_x) \leq l < 2^{g+1} f(\Delta_x)$,
then  by using similar technique of (\ref{tk2r2}), we have
\begin{eqnarray}
     P_2    \leq   \frac{1}{\exp(\frac{f(\Delta_x)\Delta_x^2}{4N^2})-1}          
            \leq  \frac{1}{\frac{n}{K} \Delta_x^2 -1}
            < \frac{K}{(1-\frac{1}{e})\Delta^2 n}
            \leq \frac{K}{(1-\frac{1}{e})\Delta_{\min}^2 n},
\end{eqnarray}
where once again we use $\frac{n \Delta_x^2}{eK} >\frac{n^{1/3} \Delta_x^2}{eK} > 1$ with $x > x_0$.

Recall that $ \widetilde{T}^2_{k}(n) \leq n \sum_{s_{1,j}\in \mathbf{s}_1}  (P_1 + P_2)  $,  thus by combining previous analysis, we have
\begin{eqnarray}
        \widetilde{T}^2_{k}(n)
                                &\leq& \frac{8 N^3 K \ln{(\frac{n}{K} N^2)}}{\Delta_{\min}^2} {n^{\frac{1}{3}}}
                                        + \frac{KN}{(1-1/e)\Delta_{\min}^2}
\label{tk2min}                 
\end{eqnarray}
Since $\Delta_{x_0} \leq \sqrt{eK/n^{1/3}}$, putting (\ref{tk1min}) and (\ref{tk2min}) in (\ref{rn_min}), we obtain
{\small\begin{eqnarray}
    \mathfrak{R}(n) &\leq&
                            \frac{e^3 K^3}{\Delta_{\min}^5}
                            +
                            NK \biggl(
                           1 + \frac{16N^2  \ln{(\frac{n^{2/3}}{K} N^2)}}{\Delta_{\min}^2}
                            + \frac{K n^{1/3}}{e^2}
                            +
                           \frac{8 N^3 K \ln{(\frac{n}{K} N^2)}}{\Delta_{\min}^2} {n^{\frac{1}{3}}}
                                        + \frac{KN}{(1-1/e)\Delta_{\min}^2}
                           \biggl)
\end{eqnarray}}

\subsection{Proof of Lemma \ref{lemma-beta-free}}
Here we still assume feasible strategy set $F$ in analysis of lower bound for all $\beta$-approximation policies. Then we adopt the same notations used in analysis of Lemma~\ref{lemma-opt-free} if not specified.
We have
\[\lambda_1 \geq \lambda_2 \geq \dots \geq \lambda_{x_\beta} \geq \dots \geq \lambda_X,\]
where $x_{\beta}$ is the greatest index of strategies satisfying $\lambda_x - \lambda_1/{\beta} \geq 0$.

Remember that we have defined $\Delta_{\beta,x}$ as the distance between $R_1/\beta$ and reward of strategy $\mathbf{s}_x$.
Similar to proof of Lemma~\ref{lemma-opt-free}, we introduce a split $x_{\beta,0}$ with $\lambda_{x_{\beta,0}} < R_1/\beta$ (or $x_{\beta,0} > x_{\beta}$)  to divide the strategies into two disjoint sets. Then the regret caused by non-$\beta$-approximation strategies can be written as
\begin{eqnarray}
   \mathfrak{R}_{\beta}(n) &\leq&    \sum \limits_{x: \lambda_x < R_1/{\beta}} \Delta_{\beta,x} E[ T_x(n)] \nonumber \\
    &\leq& n \Delta_{x_{\beta,0}} + \sum^X_{x = x_{\beta,0} +1} \Delta_{\beta,x} E[T_x(n)]  \nonumber \\
  \label{rbn}  &\leq& n \Delta_{x_{\beta,0}} + \frac{1}{\beta}\Delta_{\max} \sum^X_{x = x_{\beta,0} +1}  E[T_x(n)]
\end{eqnarray}
The last term holds due to the fact that $\frac{1}{\beta} \Delta_{\max} \geq  \Delta_{\beta,\max}$.

Using a set of counters $\{\widetilde{T}_k(n) | k=1,\dots,K\}$ to count the number of times that strategies of index $x>x_{\beta,0}$ have been played up to time slot $n$,  update $\widetilde{T}_k(n)$ if a strategy $\mathbf{s}_x$ of index $x>x_{\beta,0}$ is played for which $m_k = \min_{s_{x,j} \in \mathbf{s_x}}\{m_{s_{x,j}}\}$. we have
\begin{equation}
  \label{txtkbeta}   \sum^X_{x = x_{\beta,0} +1}  E[T_x(n)]  = \sum_{k=1}^{K} E[\widetilde{T}_k(n)].
\end{equation}

Let indicator function $I_k(t)=1$ denote the event that $\widetilde{T}_k(n)$ gets updated at time $t$, we have
\begin{eqnarray}
    \widetilde{T}_k(n) & = &      \sum_{t=1}^{n}\mathbf{ 1} \{I_k(t)=1\}                                \nonumber \\
   \label{Tk1b}                   &\leq&  \sum_{t=1}^{n} \mathbf{1}\{W_1 \leq \beta W_x(t), W_1 \geq  \beta Z_{\beta,x}  \}          \\
   \label{Tk2b}                     &&     + \sum_{t=1}^{n} \mathbf{1}\{W_1 \leq \beta W_x(t), W_1 <  \beta Z_{\beta,x} \}           \\
                      &=& \widetilde{T}^{1}_k(n) + \widetilde{T}^{2}_k(n)
\end{eqnarray}
where $\widetilde{T}^{1}_k(n)$ and $\widetilde{T}^{2}_k(n)$ respectively denote expression (\ref{Tk1b}) and (\ref{Tk2b}).

For (\ref{Tk1b}), the event $\{W_1 \leq \beta W_x(t)\}$ and $\{ W_1 \geq  \beta Z_{\beta,x}\} $  implies $\{W_x(t) \geq Z_{\beta,x}\}$.
Taking similar approaches in proof of Lemma~\ref{lemma-opt-free}, then for any positive integer $l_{\beta,0} >0$ we have
    \begin{eqnarray}
                               \widetilde{T}^{1}_k(n)  &=&  \sum_{t=1}^{n} \mathbf{1}\{W_1 \leq \beta W_x(t), W_1 \geq  \beta Z_{\beta,x}\}          \\
                            &\leq& \sum_{t=1}^{n} \mathbf{1}\{ W_x(t) \geq  Z_{\beta,x}\}                             \\
                            &\leq& \sum_{t=1}^{n} \mathbf{1}\{ W_x(t) \geq  \lambda_x + \frac{\Delta_{\beta,x}}{2} \}    \\
             &\leq& l_{\beta,0} + \sum_{t=l_{\beta,0}}^{n}
                               \mathbf{P} \biggl\{
                                    \sum_{s_{x,j} \in \mathbf{s}_x} \left(\tilde{\mu}_{s_{x,j}} +
                                    \sqrt{\frac{\ln_+( \frac{t^{2/3}}{K m_{s_{x,j}}} )} {l_{\beta,0}}} \right)                                              \geq \sum_{s_{x,j} \in \mathbf{s}_x} {\mu}_{s_{x,j}}+ \frac{\Delta_{\beta,x}}{2}, \mbox{ }\widetilde{T}^{1}_k(t)> l_{\beta,0}
                                          \biggl\}                                                                 \\
          \label{tkb1p1}   &\leq& l_{\beta,0} + \sum_{t=l_{\beta,0}}^{n}\sum_{s_{x,j} \in \mathbf{s}_x}
                                \mathbf{P} \biggl\{ \tilde{\mu}_{s_{x,j}} + \sqrt{\frac{\ln_+( \frac{t^{2/3}}{K m_{s_{x,j}}} )} {l_{\beta,0}}}                           \geq  {\mu}_{s_{x,j}}+ \frac{\Delta_{\beta,x}}{2N},  \mbox{ }\widetilde{T}^{1}_k(t)> l_{\beta,0} \biggl\}
    \end{eqnarray}
The expression of (\ref{tkb1p1}) then becomes quite the same with that of (\ref{tk1p3}) in proof of Lemma~\ref{lemma-opt-free}. By replacing $\Delta_x$ by $\Delta_{\beta,x}$, setting $l_{\beta,0}= 16N^2 \lceil \ln(\frac{n^{3/4}}{K} \Delta_{\beta,x}^2) / \Delta_{\beta,x}^2) \rceil $, and utilizing $ \Delta_{x_{\beta,0}} \geq \delta_{\beta,0} = e^{1/2} \sqrt{K/n^{2/3}}$, we have
\begin{eqnarray}
    \widetilde{T}^{1}_k(n)
     \label{tk1beta}
     &\leq& 1+ 16N^2 \frac{\ln(\frac{n^{3/4}}{K} \Delta_{\beta,x}^2)}{ \Delta_{\beta,x}^2}     +
                            K \cdot n \cdot \exp(-2\ln(n^{\frac{1}{12}}e))
                                                     \nonumber\\
                   \label{tk1br}         &= & 1+  \frac{16N^2 n^{3/4} }{K}
                   + \frac{K}{e^2}\cdot n^{\frac{5}{6}}
\end{eqnarray}
For (\ref{Tk2b}), we have
\begin{eqnarray}
    \widetilde{T}^{2}_k(n)  &\leq&  \sum_{t=1}^{n} \mathbf{1} \{W_1 \leq \beta W_x(t), W_1 <  \beta Z_{\beta,x} \} \nonumber \\
                            &\leq& n\mathbf{ P}(W_1 <  \beta Z_{\beta,x})                                                    \nonumber \\
                            &=& n \mathbf{P} \{W_1 < \lambda_1 - \frac{\beta \Delta_{\beta,x}}{2}\}                                           \nonumber\\
                            &=&  n  \mathbf{ P}   \biggl\{\sum_{s_{1,j} \in \mathbf{s}_1, j=1}^{N} w_{s_{1,j}} (z)<
                                    \lambda_1 -  \frac{\beta \Delta_{\beta,x}}{2} \biggl\}                          \nonumber\\
                             &\leq&  n  \mathbf{ P}   \biggl\{ \exists s_{1,j} \in \mathbf{s}_1: w_{s_{1,j}} (z)<  \mu_{s_{1,j}}  -
                                    \frac{\beta \Delta_{\beta,x}}{2N} \biggl\}                                       \nonumber\\
  \label{tk2br}             &\leq& n \sum_{s_{1,j} \in \mathbf{s}_1, j=1}^{N} \mathbf{ P}   \biggl\{w_{s_{1,j}} (z)<  \mu_{s_{1,j}}  -
                                    \frac{\beta \Delta_{\beta,x}}{2N} \biggl\}                                       \nonumber\\
\end{eqnarray}
To bound the value of $\mathbf{ P}   \biggl\{w_{s_{1,j}} (z)<  \mu_{s_{1,j}}-\frac{\beta \Delta_{\beta,x}}{2N} \biggl\}$,
we use function $f(u)= e \ln(\sqrt{\frac{n^{1/3}}{K}} u)/u^3$ for $u \in [\delta_{\beta,0},N]$.
Let
\begin{eqnarray}
        A_1 =    \mathbf{P}  &\biggl\{&
                        \exists 1 \leq l \leq f(\Delta_{\beta,x}):                                                                       \sum_{\tau=1}^{l} (\mu_{s_{1,j}}- \xi_{s_{1,j}}(\tau))
                        >\sqrt{l\ln_+(\frac{\tau^{2/3}}{K l})}
                \biggl\},
\end{eqnarray}
and
\begin{eqnarray}
      A_2 = \mathbf{P}    &\biggl\{&
                        \exists f(\Delta_x) < l \leq n:                                                                                      \sum_{\tau=1}^{l} (\mu_{s_{1,j}}- \xi_{s_{1,j}}(\tau))
                        > \frac{l\Delta_{\beta,x}}{2N}
                \biggl\}.
\end{eqnarray}
We have
\begin{eqnarray}
    \mathbf{ P}   \biggl\{w_{s_{1,j}} (z)<  \mu_{s_{1,j}}-\frac{\beta \Delta_{\beta,x}}{2N} \biggl\} \leq A_1+A_2.
\end{eqnarray}
Using  a peeling argument with a geometric grid of the form $\frac{1}{2^{g+1}} f(\Delta_{\beta,x}) \leq l \leq \frac{1}{2^{g}} f(\Delta_{\beta,x})$, we have
\begin{eqnarray}
        A_1 &\leq& \frac{2K f(\Delta_{\beta,x})}{n^{2/3}}
             \leq     \frac{2}{3 \sqrt{K}} n^{5/6}
\end{eqnarray}

Using a peeling argument with a geometric grid of the form $2^g f(\Delta_{\beta,x}) \leq l < 2^{g+1} f(\Delta_{\beta,x})$,
we have the following via similar technique of (\ref{tk2r2}),
\begin{eqnarray}
     A_2    &\leq&   \frac{1}{\exp(f(\Delta_{\beta,x}) \Delta_{\beta,x}^2 \beta^2 /4N^2)-1}
             \leq  \frac{4 \sqrt{K}N^2}{\beta^2}  n^{5/6}
\end{eqnarray}

Following the approaches from Equation (\ref{Tk2p1}) in proof of  Lemma~\ref{lemma-opt-free}, we can bound $\widetilde{T}^{2}_k(n)$ as:
\begin{equation}
    \widetilde{T}^{2}_k(n) \leq \frac{2 N}{3 \sqrt{K}} n^{5/6} + \frac{4 \sqrt{K}N^3}{\beta^2}  n^{5/6} \leq (1 +\frac{4 \sqrt{K}N^2}{\beta ^2}) N n^{5/6}
\end{equation}
Plugging (\ref{tk1br}), (\ref{tk2br}) into $\widetilde{T}_k(n) \leq \widetilde{T}^{1}_k(n) + \widetilde{T}^{2}_k(n) $, we have
\begin{eqnarray}
    \widetilde{T}_k(n) \leq
                         1+
                          \frac{16N^2 n^{3/4} }{K}
                         +
                            \frac{K}{e^2}\cdot n^{\frac{5}{6}}
                         +
                            (1+\frac{4 \sqrt{K}N^2}{\beta ^2})N n^{5/6}
                         \nonumber
\end{eqnarray}
With the above result and  $ \Delta_{x_{\beta,0}} \geq \delta_{\beta,0} = e^{1/2} \sqrt{K/n^{2/3}}$, we have the following bound for regret in Equation (\ref{rbn})
\begin{eqnarray}
           \mathfrak{R}_{\beta}(n)
       \leq
            N K/{\beta}
          +
            \sqrt{e K}n^{\frac{2}{3}}
          + \frac{16N^3 n^{\frac{3}{4} }}{ \beta}
          + \left(
                  1
                + \frac{4 \sqrt{K} N^2}{  \beta^2}
                +  \frac{K}{e^2N}
            \right)\frac{N^2 K}{\beta}n^{\frac{5}{6}}
\end{eqnarray}

\subsection{Proof of Lemma \ref{lemma-beta-depend}}

We then prove the results for $\beta$-approximation policy with dependency on $\Delta_{\beta,\min}$.
Without loss of generality, we still assume strategy set $F$ with $\lambda_1 \geq \dots \geq \lambda_X$.
Recall that $\Delta_{\beta,x}= \lambda_1/{\beta} - \lambda_x$, and $Z_{\beta,x} = \lambda_1/{\beta} - \Delta_{\beta,x}/2$.
Define index $x_0$  satisfying $\Delta_{\beta,x_0} \leq \delta_{\beta,0} < \Delta_{\beta, x_0+1}$ where this time we set $\delta_{\beta,0} = \sqrt{eK/n^{1/3}}$.
\begin{eqnarray}
   \mathfrak{R}_{\beta}(n) &\leq&    \sum \limits_{x: \lambda_x < R_1/{\beta}} \Delta_{\beta,x} E[ T_x(n)] \nonumber \\
    &\leq& n \Delta_{x_{\beta,0}} + \sum^X_{x = x_{\beta,0} +1} \Delta_{\beta,x} E[T_x(n)]  \nonumber \\
              &\leq& n \Delta_{x_{\beta,0}} + \frac{1}{\beta}\Delta_{\max} \sum^X_{x = x_{\beta,0} +1}  E[T_x(n)]  \nonumber \\
   \label{rbnmin}       &\leq&  n \Delta_{x_{\beta,0}} + \frac{1}{\beta}\Delta_{\max}  \sum^K_{k = 1}  E[\widetilde{T}_k(n)]
\end{eqnarray}
where the last step is from (\ref{txtkbeta}), and  $\{\widetilde{T}_k(n) | k=1,\dots,K\}$ denotes the number of times that strategies of index $x>x_{\beta,0}$ have been played up to time slot $n$.

We rewrite $\widetilde{T}_k(n) = \widetilde{T}_k^1(n) + \widetilde{T}_k^2(n)$ from (\ref{Tk1b}) and (\ref{Tk2b}), each denoting,
\begin{eqnarray}
          \widetilde{T}_k^1(n)    &=&  \sum_{t=1}^{n} \mathbf{1}\{W_1 \leq \beta W_x(t), W_1 \geq  \beta Z_{\beta,x}  \},          \\
          \widetilde{T}^{2}_k(n)  &=&      \sum_{t=1}^{n} \mathbf{1}\{W_1 \leq \beta W_x(t), W_1 <  \beta Z_{\beta,x} \}.
\end{eqnarray}
For the first term above, since $\frac{n^{1/3} \Delta_{\beta,x}^2}{K} \geq 1$ when $x > x_{\beta,0}$, we directly obtain the following from (\ref{tk1beta}),
\begin{eqnarray}
    \widetilde{T}^{1}_k(n)
                            &\leq&
     1+
     16N^2 \frac{\ln(\frac{n^{2/3}}{K} \Delta_{\beta,x}^2)}{ \Delta_{\beta,x}^2}
     +
     \sum_{t=l_0}^{n} \sum_{s_{x,j} \in \mathbf{s}_x}  \exp(-2m_{s_{x,j}}(c\Delta_{\beta,x})^2)
                            \nonumber \\
  \label{tk1bmin}             &\leq&
   1+
      16N^2 \frac{\ln(\frac{n^{2/3}}{K} \Delta_{\beta,x}^2)}{ \Delta_{\beta,x}^2}
      +
      K \cdot n \cdot \exp(-2\ln(n^{\frac{1}{3}}e))
                                            \nonumber\\
                            &\leq& 1 + \frac{16N^2  \ln{(\frac{n^{2/3}}{K} N^2)}}{\Delta_{\beta,\min}^2}
                            + \frac{K n^{1/3}}{e^2}
\end{eqnarray}
For the second term, by following (\ref{tk2br}), we have
\begin{eqnarray}
    \widetilde{T}^{2}_k(n) &\leq& n \sum_{s_{1,j} \in \mathbf{s}_1, j=1}^{N} \mathbf{ P}   \biggl\{w_{s_{1,j}} (z)<  \mu_{s_{1,j}}  -
                                    \frac{\beta \Delta_{\beta,x}}{2N} \biggl\}. \nonumber
\end{eqnarray}
And we reset the function $f(u)=4N^2 \frac{\ln{(n u^2 /K)}}{ \beta^2 u^2} $, and let
 \setlength\arraycolsep{1.5pt}
 \begin{eqnarray}
    A_1  = \mathbf{P}     & \biggl\{  &
                           \exists 1 \leq l \leq f(\Delta_{\beta,x}):
                                        \sum_{\tau=1}^{l} (\mu_{s_{1,j}}- \xi_{s_{1,j}}(\tau))
                        >\sqrt{l\ln_+(\frac{\tau^{2/3}}{K l})}
                  \biggl\},                                                                                       \\
    A_2  =  \mathbf{P}   &  \biggl\{   &
                        \exists f(\Delta_{\beta,x}) < l \leq n:
                               \sum_{\tau=1}^{l} (\mu_{s_{1,j}}- \xi_{s_{1,j}}(\tau))
                        > \frac{l\Delta_{\beta,x}}{2N}
                    \biggl\}.
 \end{eqnarray}
 We have $\mathbf{ P}   \biggl\{w_{s_{1,j}} (z)<  \mu_{s_{1,j}} -\frac{\beta \Delta_{\beta,x}}{2N} \biggl\} \leq A_1+A_2 $.

 Using  a peeling argument with a geometric grid of the form $\frac{1}{2^{g+1}} f(\Delta_{\beta,x}) \leq l \leq \frac{1}{2^{g}} f(\Delta_{\beta,x})$, we have
\begin{eqnarray}
        A_1 &\leq&      \frac{2K f(\Delta_{\beta,x})}{n^{2/3}}
             \leq       \frac{8 N^2 K}{\beta^2 n^{\frac{2}{3}}} \frac{\ln{(\frac{n}{k} \Delta_{\beta,x}^2)}}{\Delta_{\beta,x}^2}
            \leq    \frac{8 N^2 K}{\beta^2 n^{\frac{2}{3}}} \frac{\ln{(\frac{n}{k} N^2)}}{\Delta_{\beta,\min}^2}
\end{eqnarray}

Using a peeling argument with a geometric grid of the form $2^g f(\Delta_{\beta,x}) \leq l < 2^{g+1} f(\Delta_{\beta,x})$,
then we have the following by using similar technique of (\ref{tk2r2}),
\begin{eqnarray}
     A_2    &\leq&   \frac{1}{\exp(f(\Delta_{\beta,x})\Delta_{\beta,x}^2 \beta^2/4N^2)-1}
            \leq  \frac{1}{n \Delta_{\beta,x}^2 /K -1}                                    
             \leq   \frac{K}{(1-1/e)\Delta_{\beta,\min}^2 n},
\end{eqnarray}

Thus we have
\begin{eqnarray}
        \widetilde{T}^{2}_k(n) &\leq& {8 N^3 K  n^{\frac{1}{3}}} \frac{\ln{(\frac{n}{k} N^2)}}{ \beta^2 \Delta_{\beta,\min}^2}
                                    +   \frac{N K}{(1-1/e)\Delta_{\beta,\min}^2}
  \label{tk2bmin}
\end{eqnarray}

Plugging (\ref{tk1bmin}) and (\ref{tk2bmin}) in regret (\ref{rbnmin}), by $\Delta_{\beta,x_0} \leq \sqrt{eK/n^{{1}/{3}}}$, we have
\begin{eqnarray}
    \mathfrak{R}_{\beta}(n)
    &\leq&
         \frac{e^3 K^3}{\Delta_{\beta,\min}^5}
    + \frac{NK}{\beta}
    \biggl(
         1 + \frac{16N^2  \ln{(\frac{n^{2/3}}{K} N^2)}}{\Delta_{\beta,\min}^2}
                            + \frac{K n^{1/3}}{e^2}
    +
        {8 N^3 K  n^{\frac{1}{3}}} \frac{\ln{(\frac{n}{k} N^2)}}{ \beta^2 \Delta_{\beta,\min}^2}
        +   \frac{N K}{(1-1/e)\Delta_{\beta,\min}^2}
    \biggl).
    \nonumber
\end{eqnarray}

\end{document}